\documentclass[journal]{axriv}
\usepackage{amsmath,amsfonts}
\usepackage{algorithmic}
\usepackage{algorithm}
\usepackage{array}
\usepackage[caption=false,font=normalsize,labelfont=sf,textfont=sf]{subfig}
\usepackage{textcomp}
\usepackage{stfloats}
\usepackage{url}
\usepackage{verbatim}
\usepackage{graphicx}
\usepackage{cite}
\usepackage{doi}

\usepackage{arydshln}
\usepackage{ulem}
\usepackage{booktabs}
\usepackage{multirow}
\usepackage{newfloat}
\usepackage{listings}
\usepackage{epstopdf}
\usepackage{tabularx}
\usepackage{color}
\begin{document}
\let\oldtextcolor\textcolor
\renewcommand{\textcolor}[2]{#2}

\title{Dual-View Alignment Learning with Hierarchical-Prompt for Class-Imbalance Multi-Label Image Classification}

\author{Sheng Huang, Jiexuan Yan, Beiyan Liu, Bo Liu, and Richang Hong
\thanks{This work was supported in part by the National Natural Science Foundation of China under Grant 62176030 and in part by the Fundamental Research Funds for the Central Universities under Grant 2023CDJYGRH-YB18. (Corresponding author: Sheng Huang and Bo Liu)}
\thanks{Sheng Huang is with the Ministry of Education Key Laboratory of Dependable Service Computing in Cyber Physical Society, Chongqing 400044, China, and also with the School of Big Data and Software Engineering, Chongqing University, Chongqing 400044, China (e-mail: huangsheng@cqu.edu.cn).}
\thanks{Jiexuan Yan and Beiyan Liu are with the School of Big Data and Software Engineering, Chongqing University, Chongqing 400044, China (e-mail: jiexuanyan@stu.cqu.edu.cn; beiyanliu@stu.cqu.edu.cn).}
\thanks{Bo liu, and Richang Hong are with the School of Computer Science and Information Engineering, Hefei University of Technology, Hefei 230009, China (e-mail: kfliubo@gmail.com; hongrc@hfut.edu.cn).}
}

\markboth{}%
{Huang \MakeLowercase{\textit{et al.}}: Dual-View Alignment Learning with Hierarchical-Prompt for Class-Imbalance Multi-Label Classification}


\maketitle

\begin{abstract}
\textcolor{red}{
Real-world datasets often exhibit class imbalance across multiple categories, manifesting as long-tailed distributions and few-shot scenarios. This is especially challenging in Class-Imbalanced Multi-Label Image Classification (CI-MLIC) tasks, where data imbalance and multi-object recognition present significant obstacles. To address these challenges, we propose a novel method termed Dual-View Alignment Learning with Hierarchical Prompt (HP-DVAL), which leverages multi-modal knowledge from vision-language pretrained (VLP) models to mitigate the class-imbalance problem in multi-label settings. Specifically, HP-DVAL employs dual-view alignment learning to transfer the powerful feature representation capabilities from VLP models by extracting complementary features for accurate image-text alignment. To better adapt VLP models for CI-MLIC tasks, we introduce a hierarchical prompt-tuning strategy that utilizes global and local prompts to learn task-specific and context-related prior knowledge. Additionally, we design a semantic consistency loss during prompt tuning to prevent learned prompts from deviating from general knowledge embedded in VLP models. The effectiveness of our approach is validated on two CI-MLIC benchmarks: MS-COCO and VOC2007.
Extensive experimental results demonstrate the superiority of our method over SOTA approaches, achieving mAP improvements of 10.0\% and 5.2\% on the long-tailed multi-label image classification task, and 6.8\% and 2.9\% on the multi-label few-shot image classification task.}
\end{abstract}

\begin{IEEEkeywords}
Class-Imbalance Distribution, Multi-Label Classification, Image Classification, Dual-View Alignment Learning, Hierarchical Prompt.
\end{IEEEkeywords}

\section{Introduction}
In recent years, the advancements in computer vision have greatly promoted the progress of image classification \cite{He_2016_CVPR,Szegedy_2016_CVPR}. This progress greatly relies on many mainstream balanced benchmarks (e.g., CIFAR \cite{recht2018cifar10classifiersgeneralizecifar10}, MS COCO \cite{10.1007/978-3-319-10602-1_48}), which have two key characteristics: 1) they provide a relatively balanced and sufficient number of samples across all classes, and 2) each sample belongs to only one category. However, in real-world applications, the data distribution often follows a class-imbalance pattern \cite{9442377, Liu_2019_CVPR,10105457}. This class-imbalance manifests in long-tailed distributions and few-shot scenarios \cite{9207855,Yan_Zhang_Hou_Wang_Bouraoui_Jameel_Schockaert_2022,wang2024data,yan2024category}, where deep networks tend to underperform on tail or few-shot classes. Meanwhile, unlike the classical single-label classification, practical scenarios frequently involve images associated with multiple labels \cite{He_Guo_Dai_Qiao_Shu_Ren_Xia_2023, Zhu_2023_ICCV,tan2024pvlrpromptdrivenvisuallinguisticrepresentation}, adding complexity and challenge to the task. To address these issues, an increase number of works focus on the problem of Class-Imbalance Multi-Label Image Classification (CI-MLIC) \cite{10.1007/978-3-030-58548-8_10,Guo_2021_CVPR,Lin_2023, 10112637}.

\begin{figure}[t]
    \centering
    \begin{minipage}[t]{1.0\linewidth}
        \centering
        \includegraphics[width=\textwidth]{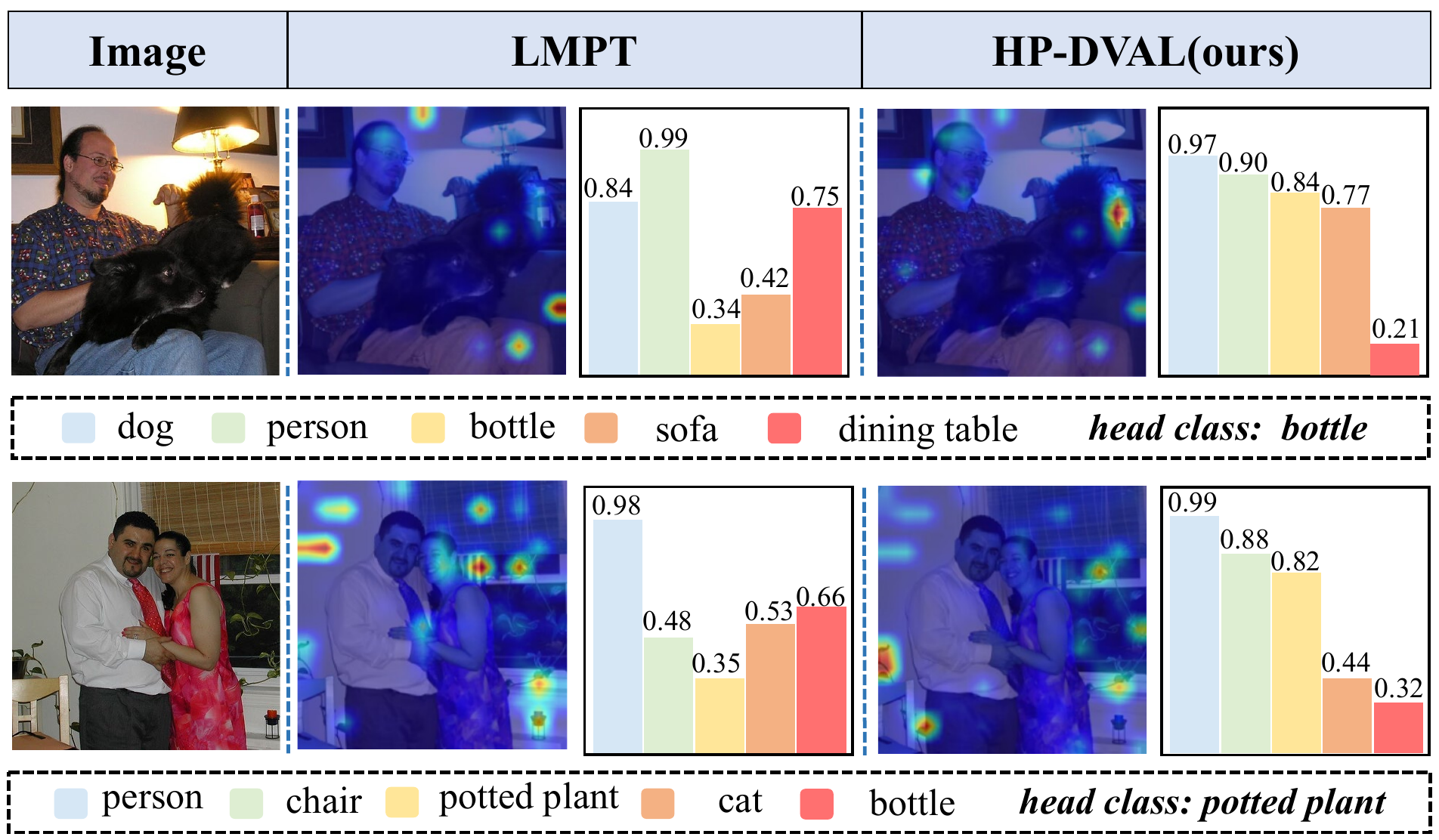}
        \centerline{(a)}
    \end{minipage}
    \begin{minipage}[t]{1.0\linewidth}
        \centering
        \includegraphics[width=\textwidth]{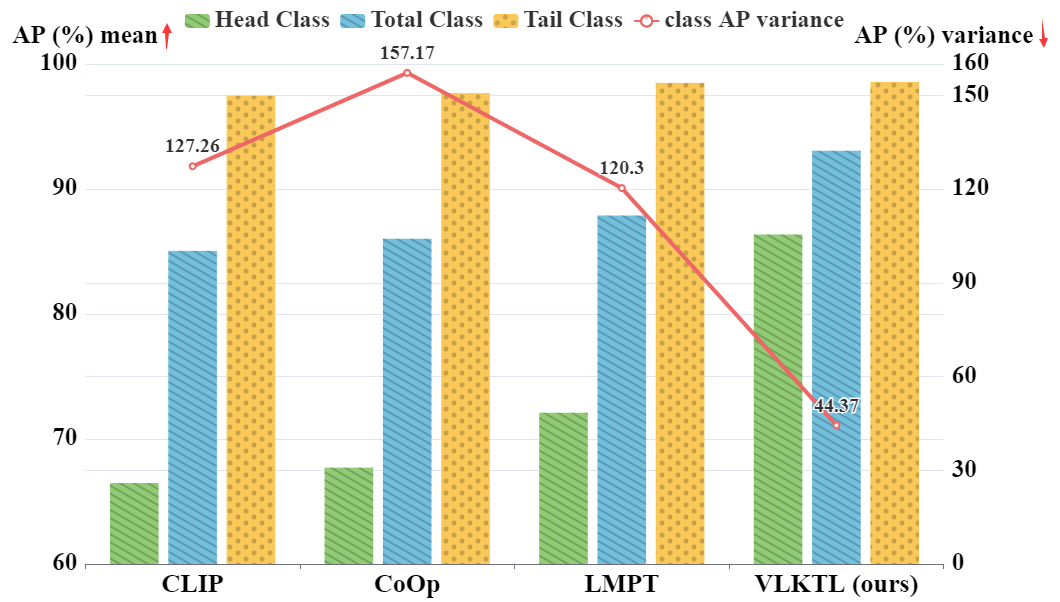}
        \centerline{(b)}
    \end{minipage}
    \caption{Comparisons of with prior prompt-tuning methods on VOC-LT dataset for multi-label long-tailed image classification task. (a) Visualization results of different methods. Prompt-tuning methods, such as LMPT \cite{xia2024lmptprompttuningclassspecific}, have poor performance for less prominent head categories (i.e., [bottle] and [potted plant]), while our HP-DVAL achieves better performance in all categories. (b) Mean and variance of prediction accuracy (AP) for different categories. Prompt-tuning methods tend to introduce significant bias towards tail classes, while our HP-DVAL can achieve balanced performance improvement on all head-to-tail category recognition.}
    \label{fig:fig 1a}
\end{figure}

\textcolor{red}{As the samples of tail or few-shot classes are relatively scarce, existing methods for solving CI-MLIC focus on addressing the sample imbalance by employing various strategies, such as resampling the number of samples for each category \cite{BUDA2018249,pmlr-v97-byrd19a}, re-weighting the loss for different categories \cite{Cui_2019_CVPR,NEURIPS2019_621461af,Guo_2021_CVPR}, and decoupling the learning of representation \cite{10382590} and classification head \cite{kang2020decouplingrepresentationclassifierlongtailed,Zhou_2020_CVPR,10284731}.} 
Recently, the rise of vision-language pretrained (VLP) models, such as CLIP \cite{pmlr-v139-radford21a}, has demonstrated the ability to learn powerful representations that effectively align image and text modalities. In these VLP models, the integration of textual modality data significantly enhances their capacity for cross-category knowledge transfer. This enhancement mitigates the adverse effects of class-imbalance distribution during the classification phase, thereby improving overall model performance. Now, these VLP models have been successfully adapted to various downstream visual tasks, particularly through prompt-tuning methods \cite{Zhou_2022_CVPR,zhou2022learning,qiu2023vtclipenhancingvisionlanguagemodels}, which offer an efficient way to transfer VLP knowledge by learning task-specific prompts rather than fine-tuning the entire model. However, when applied to class-imbalance multi-label learning, existing prompt-tuning methods tend to introduce significant bias towards tail classes, improving their performance at the expense of head classes \cite{xia2024lmptprompttuningclassspecific}, as illustrated in Figure~\ref{fig:fig 1a}. \textcolor{red}{This phenomenon can be attributed to the fact that the selected tail categories are characterized by relatively simple and easily distinguishable features (e.g., [bus] and [horse]). In contrast, for head categories such as [bottle] and [potted\_plant], the small object size and significant intra-class visual variations hinder the model’s ability to capture distinct and consistent discriminative features. Furthermore, given that CLIP’s training data is not publicly available, it is plausible that CLIP encountered samples from certain tail categories during pre-training. This pre-existing knowledge reduces the learning burden for these categories in downstream prompt-tuning.} 

\textcolor{red}{Moreover, these prompt-tuning methods frequently neglect two critical aspects, which contributes to suboptimal performance in certain categories.} Firstly, real-world images containing multiple object-categories typically require the extraction of category-specific features for accurate recognition, and the challenge of decoupling these features is further exacerbated by class-imbalance data distributions. Current prompt-tuning methods focus on the visual representation of an image from a global perspective, which makes it difficult to identify less prominent objects. \textcolor{red}{Secondly, the prompt-tuning methods rely on generic prompt templates (e.g., "a photo of a [category]") or learnable prompts, without customized designs tailored to class-imbalanced multi-label scenarios, leading to prompts that lack the necessary visual-context information of the image. In other words, these methods typically assume only a single category exists in an image, making the image-text alignment vague when the number of categories increases.} As a result, these prompts cannot accurately describe the image content, causing a mismatch between image and text representations and ultimately leading to poor performance in image-text alignment.

To address the aforementioned issues, we propose a novel image-text alignment method, termed Dual-View Alignment Learning with Hierarchical-Prompt (HP-DVAL) for CI-MLIC task. In HP-DVAL, we transfer CLIP's vision-language knowledge from both image and text modalities. From the image modality, we leverage knowledge distillation to transfer CLIP’s well-generalized visual representation capabilities to our visual encoder. To eliminate the negative impact of feature blending caused by multiple objects, we perform image-text alignment in both local and global views. Moreover, only the first \textit{k} similar image-text token pairs in the local view are selected to ensure accurate alignment. From the text modality, we propose a hierarchical prompt-tuning strategy to better adapt CLIP for class-imbalance tasks. The hierarchical prompt-tuning strategy uses a set of global and local prompts to learn hierarchical text embeddings that potentially encode task-specific and context-related prior knowledge. Additionally, we introduce a semantic consistency loss in prompt tuning to prevent learnable prompts from forgetting general knowledge encoded in CLIP. By combining dual-view alignment learning and hierarchical prompt tuning, our HP-DVAL effectively transfers CLIP's vision-language knowledge to tackle CI-MLIC tasks. 

The main technical contributions of our work are summarized as follows:

\begin{itemize}
\item We propose a novel image-text alignment method, termed Dual-View Alignment Learning with Hierarchical-Prompt (HP-DVAL) for Class-Imbalance Multi-Label Image Classification (CI-MLIC). HP-DVAL accomplishes multi-label image classification by performing image-text alignment in both local and global views. \textcolor{red}{The dual-view alignment alleviates the feature-blending issue in multi-label learning and enables the identification of less prominent objects, thereby better adapting CLIP for class-imbalanced tasks.}
\item We introduce a hierarchical prompt-tuning strategy to support dual-view image-text alignment. In this strategy, global and local text embeddings are learned separately to mine task-specific and context-related prior knowledge, thereby facilitating alignment. Additionally, we incorporate a semantic consistency loss in prompt tuning to prevent learnable prompts from forgetting the general knowledge encoded in CLIP. 
\item We conduct experiments on two CI-MLIC benchmarks, MS-COCO and Pascal VOC. Extensive results on multi-label long-tailed and few-shot image classification tasks demonstrate the significant superiority of our method over recent state-of-the-art approaches. Moreover, the experiments also highlight that our method achieves notably more balanced performance across different classes compared to other CLIP-based approaches.
\end{itemize}

\begin{figure*}[htbp]
    \centering
    \includegraphics[width=0.85\linewidth]{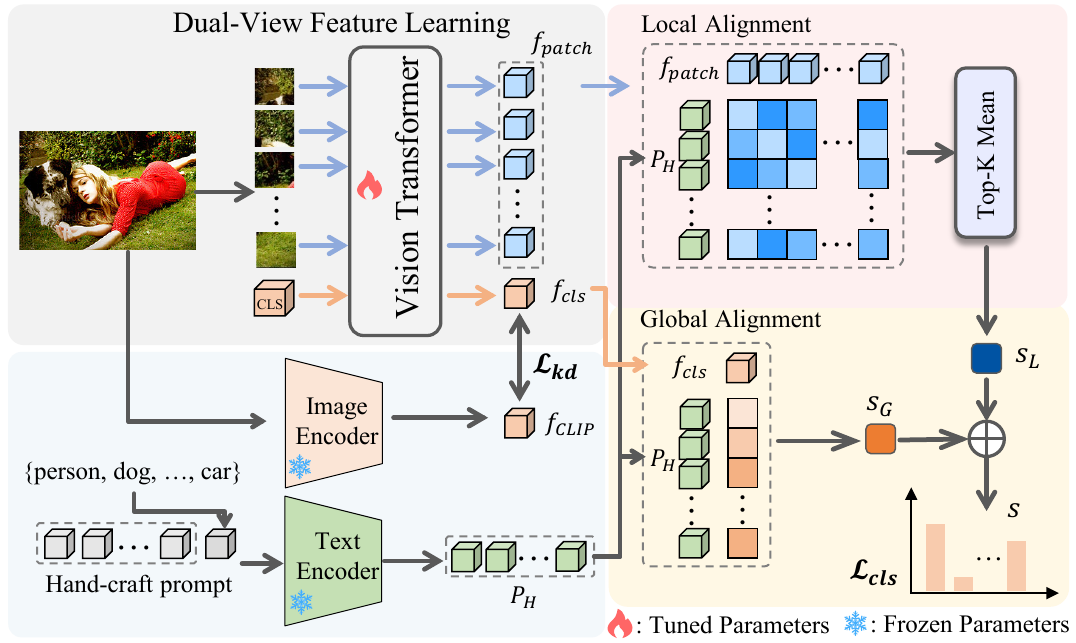}
    \caption{\textcolor{red}{
    The architecture of HP-DVAL consists of a two-stage training strategy. In the Dual-view Alignment Learning stage, we employ knowledge distillation~\cite{xu2024surveyknowledgedistillationlarge} to train a vision Transformer using CLIP's image encoder as the teacher. This stage involves dual-view feature learning to capture both global and local features. Then, we achieve alignment between the image and text embeddings from both global and local views, and integrate global and local predictions through a weighted fusion mechanism. In the Hierarchical Prompt Tuning stage, we leverage corresponding learnable global and local prompts to mine general context and task-specific knowledge relevant to the images.
    } }
    \label{fig:fig 1}
\end{figure*}

\section{Related Work}
\subsection{Class-Imbalance Learning}
Real-world data often follows a class-imbalance distribution, presenting a significant challenge for traditional models due to the rare samples of tail classes or few-shot classes. Common methods to address the long-tailed challenge include direct resampling of training samples to balance category distribution, which may result in over-fitting of the tail categories \cite{BUDA2018249,pmlr-v97-byrd19a}. Another strategy involves loss re-weighting based on label frequencies of training samples to rebalance the uneven positive gradients among classes \cite{NIPS2017_147ebe63,Cui_2019_CVPR,NEURIPS2019_621461af}. \textcolor{red}{However, these methods still lead to models learning biased features. In response, some studies \cite{10856413} have designed a diversity enhancement module to fuse information across all categories, preventing overfitting to tail classes; meanwhile, MLC-NC \cite{Tao_Li_Wan_Zheng_Chen_Li_Huang_Chen_2025} approaches the problem from the perspective of neural collapse, improving the model's representational capability by reducing intra-class differences. }
More recently, researchers have explored some techniques like transfer learning \cite{Liu_2019_CVPR,Zhu_2020_CVPR,10284731} and self-supervised learning \cite{kang2021exploring,zhang2021test,10599153} to address the long-tailed distribution. As the Vision-Language Pretrained (VLP) models like CLIP exhibit strong image-text matching ability, some strategies have been proposed to adapt them to downstream long-tailed learning tasks \cite{10.1007/978-3-031-19806-9_5,dong2023lpt,shi2024longtaillearningfoundationmodel}. These VLP-based methods incorporate additional language data to generate auxiliary confidence scores, fine-tuning the CLIP-based model on class-imbalance data. Different from these methods fine-tuning all parameters, our method leverages the powerful representation capability of CLIP to learn a feature extractor, ensuring efficient and effective adaptation to the CI-MLIC tasks.

Few-shot learning represents a distinct challenge in the context of class-imbalance distribution, as it involves predicting novel classes with rare training samples and places significant emphasis on the transfer generalization capabilities of pretrained classification networks. \textcolor{red}{ Metric learning-based methods \cite{snell2017prototypical,sung2018learning,vinyals2016matching,10230026,10106226,10938027} typically learn a metric space and classify unlabeled query samples by measuring their distances to support samples.} Meanwhile, meta learning-based methods \cite{chen2021meta,finn2017model,jamal2019task,li2021beyond,10685036} embrace a paradigm of meta-learning to enhance the few-shot adaptation capabilities of models. More recently, the Vision-Language Pretrained (VLP) models such as CLIP \cite{pmlr-v139-radford21a} exhibits strong zero-shot adaptation performance, with numerous strategies proposed to adapt it for downstream few-shot tasks \cite{Zhou_2022_CVPR,zhou2022learning,qiu2021vt,10704586}. In this paper, our method can transfer CLIP’s well-generalized visual representation capabilities across different categories to improve the performance of few-shot categories.

\subsection{Multi-Label Image Classification}

For the Multi-Label image Classification (MLC) tasks, early solutions involved training separate binary classifiers for each label \cite{tsoumakas2007multi}. Consequently, recent research addresses the MLC problem by utilizing category semantics to model label semantic correlations \cite{9600608,10663362}. CNN-based methods \cite{Wang_2016_CVPR,9466402,zhou2022learning,10122681} utilize Recurrent Neural Networks (RNNs) to extract features sensitive to label dependencies, implicitly capturing these dependencies. \textcolor{red}{With the rise of Transformer across various computer vision tasks, Transformer-based methods \cite{Lanchantin_2021_CVPR,liu2021query2labelsimpletransformerway,Nguyen_Vu_Le_2021, 10930642} utilize the core attention mechanisms in Transformer \cite{dosovitskiy2021imageworth16x16words} to explore label correlations.} However, these methods often have complex designs and may not generalize well to class-imbalance scenarios. \textcolor{red}{Recently, VLP models have been adapted for downstream MLC tasks to address few-shot and zero-shot problems \cite{Abdelfattah_2023_ICCV,He_Guo_Dai_Qiao_Shu_Ren_Xia_2023,Guo_2023_CVPR, wu2024taitextimagemultilabel,Tan_2025_CVPR,Liu_Guo_Guo_Lu_2025}.} Neglecting the imbalanced distribution of samples, they have poor generalization on long-tailed scenarios. 

In recent years, research on addressing long-tailed imbalance in multi-label settings has been relatively limited. Similar to re-weighting strategies, \cite{10.1007/978-3-030-58548-8_10} propose a distribution-balanced (DB) loss to slow down the optimization rate of negative labels based on binary-cross-entropy loss. Based on DB loss, \cite{Lin_2023} aim to flexibly adjust the training probability and further reduce the probability gap between positive and negative labels. To adapt resampling strategies to multi-label settings, \cite{Guo_2021_CVPR} adopts collaborative training on both uniform and re-balanced samplings to alleviate the class imbalance. Additionally, a prompt-tuning method \cite{xia2024lmptprompttuningclassspecific} has been proposed to adapt pretrained CLIP to long-tailed multi-label learning. However, this method may partly compromise the performance of the head classes while improving tail classes. 

On the other hand, many researchers have attempted to transfer few-shot learning methods to multi-label settings. LaSO \cite{Alfassy_2019_CVPR} was the first method to handle ML-FSIC problem with deep learning. LaSO performs intersection, union and difference set operations on class-level features, adopting a data augmentation strategy for the lack of supervision information. KGGR \cite{9207855} extends the graph structures to the few-shot domain. It introduces Graph Neural Networks (GNN) to update co-occurrence probability graphs and incorporates bilinear interpolation attention between visual embedding and label embedding to effectively model label dependencies. Meta-learning method \cite{Simon_2022_WACV} explores the way that classical few-shot methods, such as Prototypical Networks \cite{snell2017prototypical} and Relation Networks \cite{sung2018learning} expand from single-label to multi-label setting in a meta-training paradigm. These methods lack the specific designs for multi-label few-shot setting and result in poor performance. To model the label correlations between rare classes, \cite{Yan_Zhang_Hou_Wang_Bouraoui_Jameel_Schockaert_2022} explores a metric-based multi-label few-shot prototype network, using a cross-attention mechanism to extract the visual prototype corresponding to each label embedding, and measuring the distance between each visual prototype and global features. However, this cross-attention method lacks effective adjustment for prototypes. In this paper, we incorporate dual-view alignment learning and hierarchical prompt-tuning to transfer multi-modal \cite{zhang2024modality} pretrained knowledge from CLIP, achieving better performance on both head-to-tail and few-shot category recognition in CI-MLIC.

\section{Methodology}

\textbf{Overview.}
In this section, we introduce our proposed two-stage method termed Dual-View Alignment Learning with Hierarchical-Prompt (HP-DVAL), designed to tackle CI-MLIC problem by formulating it as an image-text alignment issue. This method effectively transfers pretrained knowledge across categories from CLIP, thereby alleviating the negative impacts of class imbalance.
Our method comprises two key stages, namely the dual-view alignment learning stage and the hierarchical prompt-tuning stage. These two stages will be trained sequentially for model optimization. We adopt this two-stage training strategy because simultaneously optimizing feature extractor and hierarchical prompts will lead to difficulty in convergence of learnable prompts. 

In the first stage, we utilize knowledge distillation \cite{xu2024surveyknowledgedistillationlarge} to train a Vision Transformer (ViT) \cite{dosovitskiy2021imageworth16x16words} from CLIP's image encoder, facilitating the image-text alignment, as illustrated in Figure~\ref{fig:fig 1}. To handle the presence of multiple categories within images, we adopt a dual-view feature learning approach to extract visual features and introduce image-text
alignments from both global and local views. In the second stage, similar to the visual features, we introduce two kinds of trainable prompts (i.e., global prompts and local prompts) to yield the hierarchical prompts for fine-tuning the model, as illustrated in Figure~\ref{fig:2}. 
The hierarchical prompts are elaborated to mine the general context and task-specific knowledge relevant to the images, thereby enabling CLIP to generate more accurate text embeddings for enhancing image-text alignment.
Additionally, we introduce a semantic consistency loss to prevent learnable prompts from forgetting the general knowledge encoded in CLIP.

\subsection{Dual-View Feature Learning in Vision Transformer}
In our method, the Vision Transformer (ViT) \cite{dosovitskiy2021imageworth16x16words} is used as the backbone for extracting visual features. In multi-label image classification, the global features extracted from ViT often represent a blend of visual information from multiple categories within an image. This blending can lead to the dominance of major objects in the global features, thereby suppressing the recognition of less prominent objects that coexist in the image. To address this issue, we adopt a dual-view feature learning approach within ViT, which extracts visual representations from both global and local views. The global view captures a comprehensive representation of the image, encoding the correlations between different categories, while the local view focuses on fine-grained features to achieve accurate image-text alignment.

Given an input image \(x\), we first reshape it into a sequence of flattened 2D patches \(x_p \in \mathbb{R}^{N \times (P^2 \cdot C)}\). Here \((P,P)\) denotes the resolution of each image patch, \(N\) is the number of resulting patches and \(C\) is the number of channels. We then map the patches \(x_p\) into the patch embeddings \(\bar{x}_p \in \mathbb{R}^{N \times D}\) with a linear projection, where \(D\) is the embedding dimension. The process of the \(l\)-th ViT block is formulated as:
\begin{equation}
    \begin{aligned}
    z_0 & = [x_{class}; \bar{x}_p] + E_{pos}, \\
    z'_l & = z_{l-1} + {\rm MSA}({\rm LN}(z_{l-1})),\\
    z_l & = z'_l + {\rm MLP}({\rm LN}(z'_l)),
    \end{aligned}
\end{equation}
where \(x_{class}\) represents the class token embedding and \(E_{pos}\) is the position embedding. The output of \(L\) blocks in ViT, \(z_L = [f_{cls};f_{patch}]\), consists of \(f_{cls}\) representing the class token feature (global feature) and \(f_{patch} = [f_1,f_2,...,f_N]\) representing the patch token features (local features). These features are then mapped into the embedding space using an embedding layer to align with CLIP text embeddings. The embeddings for dual-view features are denoted as \(\bar{f}_{cls}\) and \(\bar{f}_{patch} = [\bar{f}_1,\bar{f}_2,...,\bar{f}_N]\), respectively.

\subsection{Distilled Dual-View Image-Text Alignment}
Unlike CLIP's image encoder, the original ViT lacks inherent vision-language knowledge transfer ability. Therefore, we aim to equip our ViT with this ability by performing knowledge distillation \cite{xu2024surveyknowledgedistillationlarge} based on CLIP's pretrained image encoder, facilitating the subsequent image-text alignment. \textcolor{red}{The reason for not directly using the CLIP image encoder is that CLIP employs contrastive learning for category classification during training, which is inherently analogous to a standard multi-class classification task rather than a multi-label classification task. As a result, the CLIP model would inevitably allocate less attention to minor categories. Furthermore, although CLIP's original image encoder possesses vision-language alignment capabilities, its pre-training data may lack sufficient coverage of long-tailed categories in class-imbalanced datasets.} In contrast to previous prompt-tuning methods that focus solely on text prompts \cite{Zhou_2022_CVPR,zhou2022learning,qiu2023vtclipenhancingvisionlanguagemodels}, our approach leverages knowledge distillation \cite{xu2024surveyknowledgedistillationlarge} to transfer CLIP's robust knowledge from its image encoder, enhancing the alignment between the image embeddings extracted by ViT and the corresponding text embeddings. 

Specifically, we consider CLIP's image encoder \(E_{nc_{I}}\) as the teacher model and our ViT backbone as the student model. We extract CLIP image embedding as \(f_{CLIP}\), and perform the distillation on the global feature \(f_{cls}\) because both \(f_{cls}\) and \(f_{CLIP}\) are corresponded to the \([CLS]\) token. The distillation loss \(\mathcal{L}_{kd}\) is formulated to minimize the \(L_1\) distance between \(f_{CLIP}\) and \(f_{cls}\):
\begin{equation}
    \begin{aligned}
    \mathcal{L}_{kd} = ||f_{CLIP}-f_{cls}||_1, f_{CLIP} = E_{nc_{I}}(x).
    \end{aligned}
\end{equation}

During the dual-view alignment learning stage, we employ a hand-crafted template ``a photo of a [CLASS]" as input prompts to extract text embeddings from the text encoder \(E_{nc_{T}}\). These fixed prompts \(T^H =\{t^H_1,t^H_2,...,t^H_c\}\) generate text embeddings \(P^H_i\) for each class \(i\):
\begin{equation}
    P^H_i = E_{nc_{T}}(t^H_i), P^H_i \in \mathbb{R}^{1 \times D_e},
\end{equation}
where \(c\) is the number of classes and \(D_e\) is the dimension of text embeddings. In embedding space, we perform the dual-view alignment between these text embeddings \(P^H\) and the dual-view image embeddings, \(\bar{f}_{cls}\) and \(\bar{f}_{patch}\) to compute global and local prediction scores:
\begin{equation}
    \begin{aligned}
    s^G_i &= \langle P_i^H, \bar{f}_{cls} \rangle, i \in \{1,...,c\},\\
    s^L_{ij} &= \langle P^H_i, \bar{f}_j \rangle, j \in \{1,...,N\},
    \end{aligned}
\end{equation}
where \(\bar{f}_j\) is the \(j\)-th patch feature embedding and \(\langle \cdot, \cdot \rangle\) denotes the cosine similarity. The final prediction score \(s_i\) for class \(i\) integrates global and local predictions through a weighted fusion mechanism:
\begin{equation}
    s_i = \alpha s^G_i + (1-\alpha){\rm TopK}(s^L_{i1},s^L_{i2},...,s^L_{iN}),
    \label{equ 5}
\end{equation}
where \(TopK(\cdot)\) is the \textit{top-k} mean pooling, which means we focus on the \textit{top-k} most relevant local predictions, and \(\alpha \in [0,1]\) is a weight factor of dual-view alignment fusion. \textcolor{red}{The core purpose of adopting top-k mean pooling is to select the top-k high-response patches that are most relevant to the text prompts. This operation not only suppresses noise interference but also enables the model to focus on specific categories, thereby enhancing its fine-grained matching capability.}

\begin{figure}[t]
    \centering
    \includegraphics[width=0.95\linewidth]{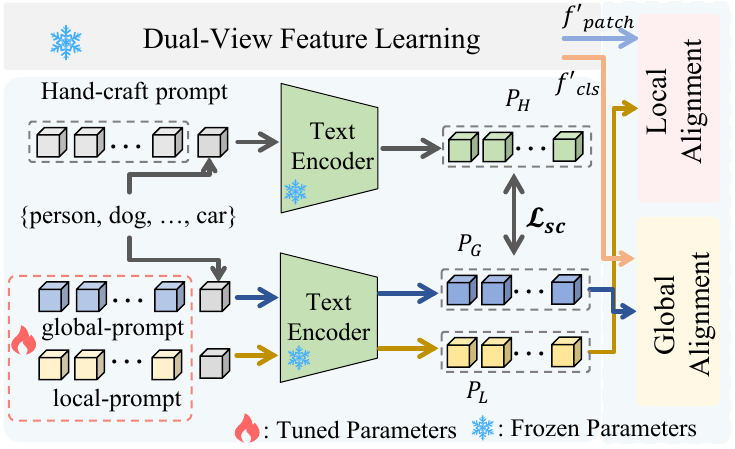}
    \caption{Hierarchical prompt-tuning stage. We introduce a series of global and local prompts to encode task-specific and context-related prior knowledge, along with semantic consistency loss to better adapt CLIP to CI-MLIC tasks.}
    \label{fig:2}
\end{figure}

\subsection{Hierarchical Prompt Tuning}
The fixed prompts based on hand-crafted templates neglect the contextual information inherent in corresponding images, which may result in suboptimal text embeddings for dual-view alignment. Therefore, we aim to learn task-specific prompts similar to previous prompt-tuning methods. In contrast to single-level prompts used in CoOp \cite{zhou2022learning}, we propose a hierarchical prompt-tuning strategy, which employs two sets of learnable prompts (global and local prompts) to derive global and local text embeddings from CLIP, as illustrated in Figure~\ref{fig:2}. The global prompts are designed to encode the comprehensive task-specific prior knowledge, while the local prompts focus on capturing the context-related prior knowledge during training.

Following the CoOp \cite{zhou2022learning}, the hierarchical prompts are formulated as:
\begin{equation}
    \begin{aligned}
    t^G_i = [v_1,v_2,...,v_M,w_i],\\
    t^L_i = [v'_1,v'_2,...,v'_M,w_i],
    \end{aligned}
\end{equation}
where \(v_k\) and \(v'_k, k \in \{1,...,M\}\) are learnable embeddings concatenated with the word embedding \(w_i\) of the \(i\)-th class to form the global prompt \(t^G_i\) and local prompt \(t^L_i\). These prompts are then fed to CLIP's text encoder \(E_{nc_{T}}\) to generate global and local text embeddings, \(P^G_i\) and \(P^L_i\):
\begin{equation}
    \begin{aligned}
    P^G_i &= E_{nc_{T}}(t^G_i), P^G_i \in \mathbb{R}^{1 \times D_e},\\
    P^L_i &= E_{nc_{T}}(t^L_i), P^L_i \in \mathbb{R}^{1 \times D_e}.
    \end{aligned}
\end{equation}

During the hierarchical prompt-tuning stage, we freeze the parameters of ViT to extract the dual-view image embeddings, \(\bar{f'}_{cls}\) and\(\bar{f'}_{patch}\) from the input image. Then we perform dual-view alignment between these learning text embeddings and the dual-view image embeddings, \(\bar{f'}_{cls}\) and \(\bar{f'}_{patch}\) to compute global and local prediction scores, respectively:
\begin{equation}
    \begin{aligned}
    s'^G_i &= \langle P_i^G, \bar{f'}_{cls} \rangle, i \in \{1,...,c\},\\
    s'^L_{ij} &= \langle P^L_i, \bar{f'}_j \rangle, j \in \{1,...,N\},
    \end{aligned}
\end{equation}
where \(\bar{f'}_j\) is the \(j\)-th patch feature embedding and \(\langle \cdot, \cdot \rangle\) denotes the cosine similarity. The final prediction score is obtained in the same weighted fusion mechanism as Equation~\ref{equ 5}:
\begin{equation}
    s'_i = \alpha s'^G_i + (1-\alpha){\rm TopK}(s'^L_{i1},s'^L_{i2},...,s'^L_{iN}),
\end{equation}
where \(TopK(\cdot)\) is the \textit{top-k} mean pooling, which means we focus on the \textit{top-k} most relevant local predictions, and \(\alpha \in [0,1]\) is a weight factor of dual-view alignment fusion.

\textcolor{red}{In the prompt initialization phase, the two hierarchical prompts share similar structural parameters, but their learning objectives diverge significantly during model optimization. Global prompt features are aligned with global visual features, emphasizing the model’s overall performance. However, because global features encompass information from multiple categories, the model often struggles to focus effectively on specific ones. In contrast, local prompt features are aligned with local visual features, focusing on image patches with high activation responses. Although these patches contain relatively limited information, their content is highly category-specific and salient, thereby enhancing the model’s ability to perform fine-grained matching. Overall, global prompts aim to capture task-specific global semantic knowledge, whereas local prompts are designed to learn semantic knowledge tied to perspective-based contexts.}

However, the learnable prompts are susceptible to overfitting rare samples and forgetting general knowledge encoded in CLIP. To mitigate this, we introduce a semantic consistency (SC) loss to enforce generalization between the learnable prompts and fixed prompts. Specifically, the SC loss \(\mathcal{L}_{sc}\) is formulated to minimize the \(L_1\) distance between \(P_{H}\) and \(P_{G}\):
\begin{equation}
    \mathcal{L}_{sc} = ||P_{H}-P_{G}||_1.
\end{equation}

\subsection{Model Optimization}
Given an input image \(x_i\), our model predicts its final category probabilities \(S_i = \{s_1^i,s_2^i,...,s_c^i\}\) and its ground truth is \(Y_i = \{y_1^i,y_2^i,...,y_c^i\}\). To address the co-occurrence of labels and the dominance of negative labels, we employ the distribution-balanced (DB) loss \cite{10.1007/978-3-030-58548-8_10} as the classification loss, which is formulated as:
\begin{equation}
    r_i = \tau + \frac{1}{1+e^{-\beta \times (\frac{1/n_i}{\sum_{i=1}^c 1/n_i} - \mu)}},
\end{equation}
\begin{equation}
    \upsilon_i = -\kappa \times -\log(\frac{1}{n_i/n}-1),
\end{equation}
\begin{equation}
    \begin{aligned}
    \mathcal{L}_{DB} = \frac{1}{c}\sum_{i=0}^c r_i [y_i\log(1+ e^{-(s_i-\upsilon_i)}) \\+ \frac{1}{\lambda}(1-y_i)\log(1+e^{\lambda(s_i-\upsilon_i)}],
    \end{aligned}
\end{equation}
where \(\tau\) is an overall lift in weight, while \(\beta\) and \(\mu\) control the shape of the mapping function. \(r_i\) is the re-balancing weight of the \(i\)-th class, \(\upsilon_i\) is the class-specific bias and \(\kappa, \lambda\) are the scale factor of DB loss. 
To further mitigate the significant decline in head-class performance caused by fine-tuning text prompts \cite{Zhou_2022_CVPR,zhou2022learning,qiu2023vtclipenhancingvisionlanguagemodels}, we incorporate class-weighting for positive samples, aimed at providing more gradient value on head classes. We modify the positive component of DB loss, and the final classification loss is defined as:
\begin{equation}
    w_i = \frac{n_i/n}{max(n_i/n)},
\end{equation}
\begin{equation}
    \begin{aligned}
    \mathcal{L}_{cls} =&\mathcal{L}_{DB\ast}= \frac{1}{c}\sum_{i=0}^c r_i [y_i\log(1+ e^{-w_i(s_i-\upsilon_i)}) 
    \\&+ \frac{1}{\lambda}(1-y_i)\log(1+e^{\lambda(s_i-\upsilon_i)}],
    \end{aligned}
\end{equation}
where \(max(\cdot)\) denotes the maximum class weight.

During model optimization, the two stages will be trained sequentially. In the dual-view alignment learning stage, the text embeddings are generated by the fixed prompts from the frozen text encoder, and the Vision Transformer is trained with the objectives of classification loss and distillation loss:
\begin{equation}
    \mathcal{L}_1 = \mathcal{L}_{cls} + \mathcal{L}_{kd}.
\end{equation}
In the hierarchical prompt-tuning stage, we freeze the parameters of the ViT and only finetune the hierarchical learnable prompts with the objectives of classification loss and semantic consistency loss:
\begin{equation}
    \mathcal{L}_2 = \mathcal{L}_{cls} + \mathcal{L}_{sc}.
\end{equation}

During the inference phrase, we utilize the trained ViT to obtain the image embeddings and use the learned hierarchical prompts to generate the text embeddings from CLIP's text encoder. Then, we compute cosine similarity between image embeddings and text embeddings from global and local views, respectively. The final classification score is obtained through a weighted fusion mechanism.

\section{Experiments and results}

\subsection{Dataset and Experimental Setup}
\textbf{1) Datasets on long-tailed multi-label image classification task.} Following the settings outlined in \cite{10.1007/978-3-030-58548-8_10}, we conduct experiments on two datasets for long-tailed multi-label image classification task: VOC-LT and COCO-LT. These two datasets are artificially sampled from two well-known multi-label image recognition benchmarks: Pascal VOC \cite{everingham2015pascal} and MS-COCO \cite{10.1007/978-3-319-10602-1_48}, respectively.
\textbf{VOC-LT} is created from the 2012 train-val set of Pascal VOC, following the guidelines provided in \cite{10.1007/978-3-030-58548-8_10}. The training set comprises 1,142 images annotated with 20 class labels. The number of images per class varies from 4 to 775. To simulate a long-tailed distribution, all classes are categorized into three groups based on the number of training samples per class: head classes (more than 100 samples), medium classes (20 to 100 samples), and tail classes (less than 20 samples). After splitting, the ratio of head, medium, and tail classes becomes 6:6:8. To evaluate the model's performance, we employ the VOC 2007 test set, which consists of 4,952 images.
Similarly, \textbf{COCO-LT} is derived from the MS-COCO 2017 dataset using a comparable approach. The training set of COCO-LT comprises 1,909 images annotated with 80 class labels. The number of images per class ranges from 6 to 1,128. The distribution of head, medium, and tail classes in COCO-LT is set to 22:33:25. The performance evaluation is conducted on the MS-COCO 2017 test set, which contains 5,000 images.

\textbf{2) Datasets on multi-label few-shot image classification task.} For multi-label few-shot image classification task, we mainly conduct experiments on MS-COCO \cite{10.1007/978-3-319-10602-1_48} and PASCAL VOC 2007 \cite{everingham2015pascal}, which are widely used benchmarks for multi-label image recognition. In order to maintain the comparability on experimental results, we used the few-shot partitioning strategy of COCO provided by LaSO~\cite{Alfassy_2019_CVPR}. Specifically, we split 80 classes to 64 base classes and 16 novel classes (\textit{bicycle, boat, stop sign, bird, backpack, frisbee, snowboard, surfboard, cup, fork, spoon, broccoli, chair, keyboard, microwave, vase}). We include images from the COCO 2014 training set, validation set and there is no overlap between the training set of \(D_{base}\) and the support set of \(D_{novel}\). For the VOC 2007 dataset, we followed ~\cite{Yan_Zhang_Hou_Wang_Bouraoui_Jameel_Schockaert_2022} to split 20 labels into 14 labels for \(C_{base}\) and and 6 labels for \(C_{novel}\) (\textit{cat, dog, pottedplant, sheep, sofa, tvmonitor}). We also sampled 10 episodes like COCO and the novel classes only appear \(K\) (\(K \in \left\{1,5\right\}\)) times in support set.

\textbf{3) Implementation Details.} 
We use the ImageNet-1K pretrained ViT-B/16 as our vision Transformer. The embedding layer for feature learning consists of two linear layers. For the VLP models, we select the pretrained CLIP with the ViT-B/16 image encoder. The patch projection of ViT-B/16 produces \(14 \times 14 = 196\) patches for an image with a resolution of \(224 \times 224\). The value of \(k\) for \textit{top-k} mean pooling is set to 32, and the weight of dual-view alignment \(\alpha\) is set to \(0.4\). For the hierarchical learnable prompts, we use the shared prompts, initializing each parameter with the Gaussian noise sampled from \(\mathcal{N}(0,0.02)\). In our experiments, both global prompts and local prompts have a sequence length of \(M = 4\), as increasing this length yields only minor improvements. Other hyperparameters in DB loss are kept the same as in \cite{10.1007/978-3-030-58548-8_10}.

In the first stage , we use the AdamW \cite{kingma2017adammethodstochasticoptimization} optimizer with a base learning  rate of \(1e-5\) and a weight decay of \(1e-4\). During the second stage, we adjust the base learning rate of the AdamW optimizer to \(5e-6\) for fine-tuning the learnable prompts. We train the model for 20 epochs with a batch size of 32 in the first stage and for 10 epochs in the second stage. All experiments are performed on one Nvidia RTX-3090 GPU, and our model is implemented in PyTorch 1.12.0.

\subsection{Evaluation Metrics}
For the long-tailed multi-label image classification task, the classes in both datasets are categorized into head, medium, and tail groups based on the number of training samples. Specifically, head classes comprise over 100 samples, medium classes consist of 20 to 100 samples each, and tail classes contain fewer than 20 samples each. We use mean average precision (\textbf{mAP}) and variance (var.) of prediction accuracy (AP) for all categories as the evaluation metric to assess the performance of long-tailed multi-label visual recognition across all classes. For the multi-label few-shot image classification task, we use mean average precision (\textbf{mAP}) on 1-shot and 5-shot settings to evaluate the performance.

To calculate mAP score, we first calculate average precision (\(AP\)) for each category \(c\) as follows:
\begin{equation}
    AP_c =\frac{ {\textstyle \sum_{n=1}^{N}}Precision\left ( n, c \right ) \cdot rel\left ( n,c \right )  }{N_c},
\end{equation}
where \(Precision\left ( n, c \right )\) is the precision for category \(c\) when retrieving \(n\) highest-ranked predicted scores and \(rel\left ( n,c \right )\) is an indicator function that is \(1\) if the image at rank \(n\) contains category \(c\) and \(0\) otherwise. \(N_c\) denotes the number of positives for category \(c\). Then \(mAP\) is computed as:
\begin{equation}
    mAP = \frac{1}{C}\sum_{c=1}^{C}AP_c,
\end{equation}
where \(C\) is the number of categories.

\begin{table*}[t]
\centering
\caption{\textcolor{red}{The mAP (\%) results of the proposed HP-DVAL and comparison methods under the long-tailed setting on two multi-label image classification datasets. Reported metrics include mAP on overall, head, medium, and tail classes, as well as the variance (var.) of average precision (AP) across all categories. Bold: best; underline: second best. $\uparrow$: higher is better; $\downarrow$: lower is better.}}
\begin{tabular}{c|ccccc|ccccc}
\hline
Datasets &
  \multicolumn{5}{c|}{VOC-LT} &
  \multicolumn{5}{c}{COCO-LT} \\ \hline
Methods &
  \multicolumn{1}{c|}{total \(\uparrow\)} &
  \multicolumn{1}{c|}{head \(\uparrow\)} &
  \multicolumn{1}{c|}{medium \(\uparrow\)} &
  \multicolumn{1}{c|}{tail \(\uparrow\)} &
  var. \(\downarrow\) &
  \multicolumn{1}{c|}{total \(\uparrow\)} &
  \multicolumn{1}{c|}{head \(\uparrow\)} &
  \multicolumn{1}{c|}{medium \(\uparrow\)} &
  \multicolumn{1}{c|}{tail \(\uparrow\)} &
  var. \(\downarrow\) \\ \hline
ERM &
  \multicolumn{1}{c|}{70.86} &
  \multicolumn{1}{c|}{68.91} &
  \multicolumn{1}{c|}{80.20} &
  \multicolumn{1}{c|}{65.31} &
  51.72 &
  \multicolumn{1}{c|}{41.27} &
  \multicolumn{1}{c|}{48.48} &
  \multicolumn{1}{c|}{49.06} &
  \multicolumn{1}{c|}{24.25} &
  129.84 \\
RW &
  \multicolumn{1}{c|}{74.70} &
  \multicolumn{1}{c|}{67.58} &
  \multicolumn{1}{c|}{82.81} &
  \multicolumn{1}{c|}{73.96} &
  47.01 &
  \multicolumn{1}{c|}{42.27} &
  \multicolumn{1}{c|}{48.62} &
  \multicolumn{1}{c|}{45.80} &
  \multicolumn{1}{c|}{32.02} &
  59.68 \\
ML-GCN \cite{Chen_2019_CVPR} &
  \multicolumn{1}{c|}{68.92} &
  \multicolumn{1}{c|}{70.14} &
  \multicolumn{1}{c|}{76.41} &
  \multicolumn{1}{c|}{62.39} &
  46.14 &
  \multicolumn{1}{c|}{44.24} &
  \multicolumn{1}{c|}{44.04} &
  \multicolumn{1}{c|}{48.36} &
  \multicolumn{1}{c|}{38.96} &
  \uline{55.92} \\
OLTR \cite{Liu_2019_CVPR} &
  \multicolumn{1}{c|}{71.02} &
  \multicolumn{1}{c|}{70.31} &
  \multicolumn{1}{c|}{79.80} &
  \multicolumn{1}{c|}{64.95} &
  50.02 &
  \multicolumn{1}{c|}{45.83} &
  \multicolumn{1}{c|}{47.45} &
  \multicolumn{1}{c|}{50.63} &
  \multicolumn{1}{c|}{38.05} &
  59.51 \\
LDAM \cite{NEURIPS2019_621461af} &
  \multicolumn{1}{c|}{70.73} &
  \multicolumn{1}{c|}{68.73} &
  \multicolumn{1}{c|}{80.38} &
  \multicolumn{1}{c|}{69.09} &
  48.96 &
  \multicolumn{1}{c|}{40.53} &
  \multicolumn{1}{c|}{48.77} &
  \multicolumn{1}{c|}{48.38} &
  \multicolumn{1}{c|}{22.92} &
  142.79 \\
CB Focal \cite{Cui_2019_CVPR} &
  \multicolumn{1}{c|}{75.24} &
  \multicolumn{1}{c|}{70.30} &
  \multicolumn{1}{c|}{83.53} &
  \multicolumn{1}{c|}{72.74} &
  52.04 &
  \multicolumn{1}{c|}{49.06} &
  \multicolumn{1}{c|}{47.91} &
  \multicolumn{1}{c|}{53.01} &
  \multicolumn{1}{c|}{44.85} &
  62.49 \\
BBN \cite{Zhou_2020_CVPR} &
  \multicolumn{1}{c|}{73.37} &
  \multicolumn{1}{c|}{71.31} &
  \multicolumn{1}{c|}{81.76} &
  \multicolumn{1}{c|}{68.62} &
  53.07 &
  \multicolumn{1}{c|}{50.00} &
  \multicolumn{1}{c|}{49.79} &
  \multicolumn{1}{c|}{53.99} &
  \multicolumn{1}{c|}{44.91} &
  64.86 \\
DB Focal \cite{10.1007/978-3-030-58548-8_10} &
  \multicolumn{1}{c|}{78.94} &
  \multicolumn{1}{c|}{73.22} &
  \multicolumn{1}{c|}{84.18} &
  \multicolumn{1}{c|}{79.30} &
  49.06 &
  \multicolumn{1}{c|}{53.55} &
  \multicolumn{1}{c|}{51.13} &
  \multicolumn{1}{c|}{57.05} &
  \multicolumn{1}{c|}{51.06} &
  58.71 \\
ASL \cite{Ridnik_2021_ICCV} &
  \multicolumn{1}{c|}{76.40} &
  \multicolumn{1}{c|}{70.70} &
  \multicolumn{1}{c|}{82.26} &
  \multicolumn{1}{c|}{76.29} &
  51.11 &
  \multicolumn{1}{c|}{50.21} &
  \multicolumn{1}{c|}{49.05} &
  \multicolumn{1}{c|}{53.65} &
  \multicolumn{1}{c|}{46.68} &
  59.26 \\
LTML \cite{Guo_2021_CVPR} &
  \multicolumn{1}{c|}{81.44} &
  \multicolumn{1}{c|}{75.68} &
  \multicolumn{1}{c|}{85.53} &
  \multicolumn{1}{c|}{82.69} &
  46.42 &
  \multicolumn{1}{c|}{56.90} &
  \multicolumn{1}{c|}{54.13} &
  \multicolumn{1}{c|}{60.59} &
  \multicolumn{1}{c|}{54.47} &
  61.55 \\
Stitch-Up \cite{10112637} &
  \multicolumn{1}{c|}{76.48} &
  \multicolumn{1}{c|}{67.85} &
  \multicolumn{1}{c|}{80.87} &
  \multicolumn{1}{c|}{79.67} &
  53.89 &
  \multicolumn{1}{c|}{54.14} &
  \multicolumn{1}{c|}{\uline{59.80}} &
  \multicolumn{1}{c|}{51.53} &
  \multicolumn{1}{c|}{51.27} &
  64.20 \\
CDRS+AFL \cite{10332128} &
  \multicolumn{1}{c|}{78.96} &
  \multicolumn{1}{c|}{73.35} &
  \multicolumn{1}{c|}{85.03} &
  \multicolumn{1}{c|}{78.63} &
  51.62 &
  \multicolumn{1}{c|}{55.35} &
  \multicolumn{1}{c|}{52.45} &
  \multicolumn{1}{c|}{59.48} &
  \multicolumn{1}{c|}{52.46} &
  62.11 \\
Bilateral-TPS \cite{10394132} &
  \multicolumn{1}{c|}{81.58} &
  \multicolumn{1}{c|}{\uline{75.88}} &
  \multicolumn{1}{c|}{84.11} &
  \multicolumn{1}{c|}{83.95} &
  \uline{44.65} &
  \multicolumn{1}{c|}{56.38} &
  \multicolumn{1}{c|}{55.93} &
  \multicolumn{1}{c|}{58.26} &
  \multicolumn{1}{c|}{54.29} &
  62.92 \\
CAE-Net \cite{10351721} &
  \multicolumn{1}{c|}{81.61} &
  \multicolumn{1}{c|}{74.00} &
  \multicolumn{1}{c|}{85.35} &
  \multicolumn{1}{c|}{85.28} &
  48.28 &
  \multicolumn{1}{c|}{57.64} &
  \multicolumn{1}{c|}{52.37} &
  \multicolumn{1}{c|}{61.18} &
  \multicolumn{1}{c|}{57.63} &
  62.97 \\
PG Loss \cite{Lin_2023} &
  \multicolumn{1}{c|}{80.37} &
  \multicolumn{1}{c|}{73.67} &
  \multicolumn{1}{c|}{83.83} &
  \multicolumn{1}{c|}{82.88} &
  50.61 &
  \multicolumn{1}{c|}{54.43} &
  \multicolumn{1}{c|}{51.23} &
  \multicolumn{1}{c|}{57.42} &
  \multicolumn{1}{c|}{53.40} &
  56.92 \\
COMIC \cite{Zhang_2023_ICCV} &
  \multicolumn{1}{c|}{81.53} &
  \multicolumn{1}{c|}{73.10} &
  \multicolumn{1}{c|}{89.18} &
  \multicolumn{1}{c|}{84.53} &
  53.73 &
  \multicolumn{1}{c|}{55.08} &
  \multicolumn{1}{c|}{49.21} &
  \multicolumn{1}{c|}{60.08} &
  \multicolumn{1}{c|}{55.36} &
  59.77 \\ \hline
\textit{CLIP:ViT16} &
  \multicolumn{1}{c|}{} &
  \multicolumn{1}{c|}{} &
  \multicolumn{1}{c|}{} &
  \multicolumn{1}{c|}{} &
   &
  \multicolumn{1}{c|}{} &
  \multicolumn{1}{c|}{} &
  \multicolumn{1}{c|}{} &
  \multicolumn{1}{c|}{} &
   \\
Zero-Shot CLIP &
  \multicolumn{1}{c|}{85.05} &
  \multicolumn{1}{c|}{66.48} &
  \multicolumn{1}{c|}{87.08} &
  \multicolumn{1}{c|}{97.46} &
  127.26 &
  \multicolumn{1}{c|}{60.17} &
  \multicolumn{1}{c|}{38.52} &
  \multicolumn{1}{c|}{65.06} &
  \multicolumn{1}{c|}{72.28} &
  184.57 \\
CoOp \cite{zhou2022learning} &
  \multicolumn{1}{c|}{86.02} &
  \multicolumn{1}{c|}{67.71} &
  \multicolumn{1}{c|}{88.79} &
  \multicolumn{1}{c|}{97.67} &
  157.17 &
  \multicolumn{1}{c|}{60.68} &
  \multicolumn{1}{c|}{41.97} &
  \multicolumn{1}{c|}{63.18} &
  \multicolumn{1}{c|}{73.85} &
  153.05 \\
CoCoOp \cite{Zhou_2022_CVPR} &
  \multicolumn{1}{c|}{84.47} &
  \multicolumn{1}{c|}{64.58} &
  \multicolumn{1}{c|}{87.82} &
  \multicolumn{1}{c|}{96.88} &
  182.46 &
  \multicolumn{1}{c|}{61.49} &
  \multicolumn{1}{c|}{39.81} &
  \multicolumn{1}{c|}{64.63} &
  \multicolumn{1}{c|}{76.42} &
  202.98 \\
LMPT \cite{xia2024lmptprompttuningclassspecific} &
  \multicolumn{1}{c|}{\uline{87.88}} &
  \multicolumn{1}{c|}{72.10} &
  \multicolumn{1}{c|}{\uline{89.26}} &
  \multicolumn{1}{c|}{\uline{98.49}}&
  120.30 &
  \multicolumn{1}{c|}{\uline{66.19}} &
  \multicolumn{1}{c|}{44.89} &
  \multicolumn{1}{c|}{\uline{69.80}} &
  \multicolumn{1}{c|}{\uline{79.08}} &
  181.95 \\
HP-DVAL(ours) &
  \multicolumn{1}{c|}{\textbf{93.06}} &
  \multicolumn{1}{c|}{\textbf{86.36}} &
  \multicolumn{1}{c|}{\textbf{92.83}} &
  \multicolumn{1}{c|}{\textbf{98.58}} &
  \textbf{44.37} &
  \multicolumn{1}{c|}{\textbf{76.19}} &
  \multicolumn{1}{c|}{\textbf{72.05}} &
  \multicolumn{1}{c|}{\textbf{75.98}} &
  \multicolumn{1}{c|}{\textbf{79.12}} &
  \textbf{54.91} \\ \hline
\end{tabular}%
\label{tab 1}
\end{table*}

\begin{table}[t]
\centering
\caption{ \textcolor{red}{The mAP (\%) comparison of the proposed method with SOTA multi-label few-shot image classification methods on the VOC2007 and MS-COCO datasets. Best: bold; second best: underlined.}}
\begin{tabular}{l|cc|cc}
\hline
                     & \multicolumn{2}{c|}{VOC2007} & \multicolumn{2}{c}{MS-COCO}   \\ \cline{2-5} 
 \multirow{-2}{*}{Methods} & 1-shot        & 5-shot       & 1-shot        & 5-shot        \\ \hline
LaSO \cite{Alfassy_2019_CVPR} &     47.9          &    59.1          & 45.3          & 58.1          \\
ML-FSL \cite{Simon_2022_WACV}                  &    50.2           &   60.4           & 54.4          & 63.6          \\
KGGR \cite{9207855}                    &      49.4         &     61.0         & 52.3          & 63.5          \\
NLC \cite{Simon_2022_WACV}                     &      53.3         &     60.8         & 56.8          & 64.8          \\
WAGP \cite{Yan_Zhang_Hou_Wang_Bouraoui_Jameel_Schockaert_2022}                    &     53.3          &     69.3         & 55.7          & 68.2          \\
BCR \cite{10508813}                     &     59.8          &    66.7          & 63.7          & 72.2          \\
LCM \cite{yan2024modellingmultimodalcrossinteractionmlfsic}                     &    62.9           &    71.7          & 64.2          & \uline{73.9}          \\ \hline
\textit{CLIP:ViT16}      &               &              &               &               \\
CoOp \cite{zhou2022learning}                    & 79.3          & 83.8         & 52.6          & 58.1          \\
TaI-DPT \cite{Guo_2023_CVPR}                 & 83.2          & 88.2         & 65.8          & 67.6          \\
CoMC \cite{liulanguage}                     & 89.7          & 90.6         & 68.9          & 70.4          \\
PVP  \cite{wu2024taitextimagemultilabel}                    & \uline{90.1}          & \uline{92.5}         & \uline{70.9}          & 72.4 \\
HP-DVAL(ours)            &  \textbf{90.8}             &  \textbf{95.4}            & \textbf{73.5} & \textbf{79.2} \\ \hline
\end{tabular}
\label{tab 2}
\end{table}

\subsection{Experimental Results}
\textbf{1) Results on long-tailed multi-label image classification task.}
To validate the effectiveness of our proposed method, we compare it with previous state-of-the-art methods on long-tailed multi-label image classification task. These methods include Empirical Risk Minimization (ERM), a smooth version of Re-Weighting (RW) using the inverse proportion to the square root of class frequency, ML-GCN \cite{Chen_2019_CVPR}, OLTR \cite{Liu_2019_CVPR}, LDAM \cite{NEURIPS2019_621461af}, Class-Balanced (CB) Focal \cite{Cui_2019_CVPR}, BBN \cite{Zhou_2020_CVPR}, Distribution-Balanced (DB) Focal \cite{10.1007/978-3-030-58548-8_10}, ASL \cite{Ridnik_2021_ICCV}, LTML \cite{Guo_2021_CVPR}, Stitch-Up \cite{10112637}, CDRS+AFL \cite{10332128}, Bilateral-TPS \cite{10394132}, CAE-Net \cite{10351721}, PG Loss \cite{Lin_2023} and COMIC \cite{Zhang_2023_ICCV}. We also compare our HP-DVAL with previous prompt-tuning methods, including CoOp \cite{zhou2022learning}, CoCoOp \cite{Zhou_2022_CVPR} and LMPT \cite{xia2024lmptprompttuningclassspecific}. Following the settings outlined in \cite{10.1007/978-3-030-58548-8_10}, we conduct experiments on two long-tailed datasets: VOC-LT and COCO-LT. These two datasets are artificially sampled from VOC and MS-COCO, respectively. The experimental results on the long-tailed multi-label image classification task are shown in Table~\ref{tab 1}. Experimental results demonstrate that our proposed method significantly outperforms previous methods. Specifically, our proposed HP-DVAL achieves outstanding results, with total mAP of 93.06\% on VOC-LT and 76.19\% on COCO-LT. Compared to the previous SOTA method (LMPT), HP-DVAL shows notable improvements of approximately 5.18\% and 10.0\% in total mAP on VOC-LT and COCO-LT, repsectively. Moreover, HP-DVAL excels across all three class subsets (head, medium, and tail classes), with particularly noteworthy gains in head classes, achieving a remarkable 14.26\% and 27.16\% mAP increase on COCO-LT and VOC-LT over the LMPT method. Compared with previous prompt-tuning methods that are biased towards tail classes, our method significantly enhances the head-class performance while also improving tail-class performance. We attribute this to the fact that previous prompt-tuning methods fail to design
task-specific prompts that account for long-tailed distributions, leading to prompts that lack the necessary visual-context information of the image. Moreover, our method achieves the minimum variance of prediction accuracy for all categories, which means our HP-DVAL can achieve balanced performance improvement on all head-to-tail category recognition. These results fully demonstrate the efficacy of HP-DVAL in achieving synchronous improvements in head-to-tail category performance for long-tailed multi-label learning.

\begin{table}[t]
    \caption{\textcolor{red}{The mAP (\%) results of the proposed HP-DVAL under different multi-label image classification losses on the COCO-LT dataset. $\mathrm{DB}^*$ denotes the proposed classification loss.}}
   \label{tab 3}
   \centering
\begin{tabular}{ccccc}
\hline
\multicolumn{1}{c|}{Dataset}        & \multicolumn{4}{c}{COCO-LT}                                                                                   \\ \hline
\multicolumn{1}{c|}{Loss Functions} & \multicolumn{1}{c|}{total} & \multicolumn{1}{c|}{head}  & \multicolumn{1}{c|}{medium} & tail                 \\ \hline
\multicolumn{1}{c|}{BCE}            & \multicolumn{1}{c|}{69.58} & \multicolumn{1}{c|}{66.71} & \multicolumn{1}{c|}{69.46}  & 71.59                \\
\multicolumn{1}{c|}{MLS}            & \multicolumn{1}{c|}{71.85} & \multicolumn{1}{c|}{68.33} & \multicolumn{1}{c|}{71.53}  & 74.46                \\
\multicolumn{1}{c|}{CB Focal}       & \multicolumn{1}{c|}{72.39} & \multicolumn{1}{c|}{68.80} & \multicolumn{1}{c|}{72.26}  & 74.89                \\
\multicolumn{1}{c|}{ASL}    & \multicolumn{1}{c|}{74.23} & \multicolumn{1}{c|}{\uline{69.85}} & \multicolumn{1}{c|}{73.92}  & 77.39                \\
\multicolumn{1}{c|}{DB Focal}       & \multicolumn{1}{c|}{\uline{75.16}} & \multicolumn{1}{c|}{69.72} & \multicolumn{1}{c|}{\uline{75.52}}  & \uline{78.50}                \\
\multicolumn{1}{c|}{$\mathrm{DB}^*$(ours)}            & \multicolumn{1}{c|}{\textbf{76.19}} & \multicolumn{1}{c|}{\textbf{72.05}} & \multicolumn{1}{c|}{\textbf{75.98}}  & \textbf{79.12}                \\ \hline
\end{tabular}
\end{table}

\begin{table}[t]
\caption{\textcolor{red}{The mAP (\%) results of different methods under BCE loss and DB loss on the COCO-LT dataset. Best: bold; second best: underlined.}}
   \label{tab 4}
\centering
\begin{tabular}{l|cc|cccc}
\hline
\multirow{2}{*}{Methods} & \multicolumn{1}{c|}{\multirow{2}{*}{BCE}} & \multirow{2}{*}{DB} & \multicolumn{4}{c}{COCO-LT}                                                                                                     \\ \cline{4-7} 
                         & \multicolumn{1}{c|}{}                     &                             & \multicolumn{1}{c|}{total}          & \multicolumn{1}{c|}{head}           & \multicolumn{1}{c|}{medium}         & tail           \\ \hline
BCE                      & -                                         & -                           & \multicolumn{1}{c|}{49.43}          & \multicolumn{1}{c|}{48.77}          & \multicolumn{1}{c|}{53.00}          & 45.33          \\
DB Focal                & -                                         & -                           & \multicolumn{1}{c|}{53.55}          & \multicolumn{1}{c|}{51.13}          & \multicolumn{1}{c|}{57.05}          & 51.06          \\ \hline
\multirow{2}{*}{CoOp}    & \(\surd\)                                 & -                           & \multicolumn{1}{c|}{55.03}          & \multicolumn{1}{c|}{39.80}          & \multicolumn{1}{c|}{57.71}          & 64.91          \\
                         & -                                         & \(\surd\)                   & \multicolumn{1}{c|}{60.68}          & \multicolumn{1}{c|}{41.97}          & \multicolumn{1}{c|}{63.18}          & 73.85          \\
\multirow{2}{*}{CoCoOp}  & \(\surd\)                                 & -                           & \multicolumn{1}{c|}{58.81}          & \multicolumn{1}{c|}{45.49}          & \multicolumn{1}{c|}{61.78}          & 66.61          \\
                         & -                                         & \(\surd\)                   & \multicolumn{1}{c|}{61.49}          & \multicolumn{1}{c|}{39.81}          & \multicolumn{1}{c|}{64.63}          & 76.42          \\
\multirow{2}{*}{LMPT}    & \(\surd\)                                 & -                           & \multicolumn{1}{c|}{58.04}          & \multicolumn{1}{c|}{41.79}          & \multicolumn{1}{c|}{58.86}          & 73.90          \\
                         & -                                         & \(\surd\)                   & \multicolumn{1}{c|}{66.19}          & \multicolumn{1}{c|}{44.89}          & \multicolumn{1}{c|}{\uline{69.80}}          & \textbf{79.08}          \\
\multirow{2}{*}{HP-DVAL} & \(\surd\)                                 & -                           & \multicolumn{1}{c|}{\uline{69.58}}          & \multicolumn{1}{c|}{\uline{66.71}}          & \multicolumn{1}{c|}{69.46}          & 71.59          \\
                         & -                                         & \(\surd\)                   & \multicolumn{1}{c|}{\textbf{75.16}} & \multicolumn{1}{c|}{\textbf{69.72}} & \multicolumn{1}{c|}{\textbf{75.52}} & \uline{78.50} \\ \hline
\end{tabular}
\end{table}

\begin{table*}[t]
\caption{\textcolor{red}{Ablation study on the components of the proposed HP-DVAL, where DVAL = Dual-View Alignment Learning, HPT = Hierarchical Prompt Tuning, and SC = Semantic Consistency. Best scores are shown in bold.}}
\label{tab 5}
\centering
\begin{tabular}{c|ccc|cccc|cccc}
\hline
\multirow{2}{*}{} &
  \multicolumn{1}{c|}{\multirow{2}{*}{DVAL}} &
  \multicolumn{1}{c|}{\multirow{2}{*}{HPT}} &
  \multirow{2}{*}{SC} &
  \multicolumn{4}{c|}{VOC-LT} &
  \multicolumn{4}{c}{COCO-LT} \\ \cline{5-12} 
 &
  \multicolumn{1}{c|}{} &
  \multicolumn{1}{c|}{} &
   &
  \multicolumn{1}{c|}{total} &
  \multicolumn{1}{c|}{head} &
  \multicolumn{1}{c|}{medium} &
  tail &
  \multicolumn{1}{c|}{total} &
  \multicolumn{1}{c|}{head} &
  \multicolumn{1}{c|}{medium} &
  tail \\ \hline
 &
   &
   &
   &
  \multicolumn{1}{c|}{85.05} &
  \multicolumn{1}{c|}{66.48} &
  \multicolumn{1}{c|}{87.08} &
  97.46 &
  \multicolumn{1}{c|}{60.17} &
  \multicolumn{1}{c|}{38.52} &
  \multicolumn{1}{c|}{65.06} &
  72.28 \\
Two-stage &
  \(\surd\) &
   &
   &
  \multicolumn{1}{c|}{91.59} &
  \multicolumn{1}{c|}{83.84} &
  \multicolumn{1}{c|}{91.59} &
  96.42 &
  \multicolumn{1}{c|}{75.72} &
  \multicolumn{1}{c|}{71.76} &
  \multicolumn{1}{c|}{75.39} &
  78.62 \\
 &
  \(\surd\) &
  \(\surd\) &
   &
  \multicolumn{1}{c|}{92.94} &
  \multicolumn{1}{c|}{86.28} &
  \multicolumn{1}{c|}{92.70} &
  98.44 &
  \multicolumn{1}{c|}{76.02} &
  \multicolumn{1}{c|}{71.76} &
  \multicolumn{1}{c|}{75.96} &
  78.88 \\
 &
  \(\surd\) &
  \(\surd\) &
  \(\surd\) &
  \multicolumn{1}{c|}{\textbf{93.06}} &
  \multicolumn{1}{c|}{\textbf{86.36}} &
  \multicolumn{1}{c|}{\textbf{92.83}} &
  \textbf{98.58} &
  \multicolumn{1}{c|}{\textbf{76.19}} &
  \multicolumn{1}{c|}{\textbf{72.05}} &
  \multicolumn{1}{c|}{\textbf{75.98}} &
  \textbf{79.12} \\ \hline
Joint &
  \(\surd\) &
  \(\surd\) &
  \(\surd\) &
  \multicolumn{1}{c|}{90.51} &
  \multicolumn{1}{c|}{93.34} &
  \multicolumn{1}{c|}{92.00} &
  \multicolumn{1}{c|}{94.62} &
  \multicolumn{1}{c|}{73.36} &
  \multicolumn{1}{c|}{69.45} &
  \multicolumn{1}{c|}{75.04} &
  \multicolumn{1}{c}{74.25} \\ \hline
\end{tabular}%
\end{table*}

\textbf{2) Results on multi-label few-shot image classification task.}
For the multi-label few-shot task, we compare our HP-DVAL with previous few-shot methods, which include LaSO \cite{Alfassy_2019_CVPR}, ML-FSL \cite{Simon_2022_WACV}, KGGR \cite{9207855}, WAGP \cite{Yan_Zhang_Hou_Wang_Bouraoui_Jameel_Schockaert_2022}, BCR \cite{10508813} and LCM \cite{Yan_Zhang_Hou_Wang_Bouraoui_Jameel_Schockaert_2022}. 
We also evaluate our method against CLIP-based methods that only use text data for training and integrate with CoOp \cite{zhou2022learning}, i.e., TAI-DPT \cite{Guo_2023_CVPR}, CoMC \cite{liulanguage} and PVP \cite{wu2024taitextimagemultilabel}. 
The experimental results for multi-label few-shot learning on MS-COCO and VOC 2007 dataset are shown in Table~\ref{tab 2}. From Table~\ref{tab 2}, it is evident that our HP-DVAL outperforms the SOTA few-shot method by a large margin, achieving improvements of 9.3\%, 5.3\% mAP on 1-shot and 5-shot settings on MS-COCO and 27.9\%, 23.7\% mAP on 1-shot and 5-shot settings on MS-COCO, respectively. We attribute the performance improvement to our method's ability to effectively enhance CLIP's capability for cross-category knowledge transfer.
Furthermore, compared to the CLIP-based methods, our HP-DVAL still has certain performance superiority on both datasets, with a maximum improvement of 6.8\% mAP on 5-shot, MS-COCO.
These results fully demonstrate the efficacy of HP-DVAL in achieving substantial improvements in few-shot category recognition for class-imbalance learning.

\begin{figure}[t]
\centering
    \begin{minipage}[t]{0.72\linewidth}
        \centering
        \includegraphics[width=\textwidth]{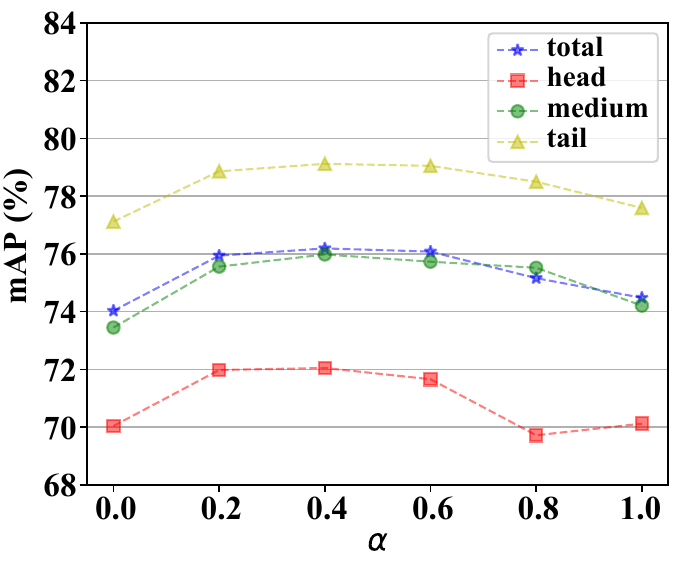}
        \centerline{(a) Effect of \(\alpha\)}
    \end{minipage}
    \begin{minipage}[t]{0.72\linewidth}
        \centering
        \includegraphics[width=\textwidth]{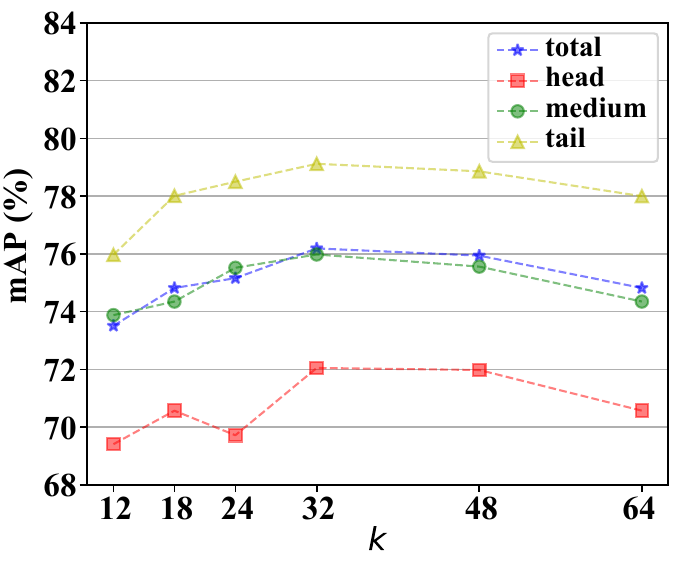}
        \centerline{(b) Effect of \(k\)}
    \end{minipage}
    \caption{\textcolor{red}{Impact of hyper-parameters.The mAP (\%) results for different dual-view alignment weight \(\alpha\) and different \(k\) in \textit{top-k} mean pooling on COCO-LT dataset.}}
    \label{fig:fig 2}
    \vspace{-0.3cm}
\end{figure}
\subsection{Ablation Studies}

\textbf{1) Multi-Label Classification Loss.}
To mitigate the significant decline in head-class performance caused by fine-tuning prompts, we modify the positive component of original Distribution-Balanced loss. In this part, we compare several classification losses for optimizing our HP-DVAL method, including Binary Cross-Entropy Loss (BCE), Multi-Label Soft Margin Loss (MSL), Class-Balanced Focal Loss (CB Focal) \cite{Cui_2019_CVPR}, Asymmetric Loss (ASL) \cite{Ridnik_2021_ICCV}, DB Focal \cite{10.1007/978-3-030-58548-8_10} and our classification loss (denoted as $\mathrm{DB}^*$). Table~\ref{tab 3} lists the comparison results on COCO-LT. The results demonstrate that our $\mathrm{DB}^*$ consistently outperforms other classification losses for the CI-MLIC tasks. Compared to the original DB loss, our $\mathrm{DB}^*$ significantly improves head-class performance, achieving an increase of 2.33\%. This superior performance can be attributed to the ability to incorporate class-weighting for positive samples and address the dominance of negative labels, a key aspect that other losses do not adequately address.

To further verify that the performance improvement achieved by our method does not come from the advantage of DB loss, we compare the experimental results of different methods on BCE loss and DB loss \cite{10.1007/978-3-030-58548-8_10}. Table~\ref{tab 4} lists the comparison results on COCO-LT. The results demonstrate that our method outperforms all previous methods on overall and head class mAP without using DB loss, especially achieving an increase of 21.22\% on head mAP. Integrating DB loss into our method can further improve the performance of tail classes, thereby enhancing the prediction of overall class distribution. These results fully demonstrate that our HP-DVAL does not rely on DB loss to achieve synchronous improvement in head-to-tail category performance, but rather that DB loss can help to improve better tail-class performance.

\textbf{2) Sensitivity of Hyper-Parameters.}
We further explore the impact of the dual-view alignment weight \(\alpha\) and the \(k\) parameter in \textit{top-k} pooling on the performance of our method on the COCO-LT dataset. The mAP results are presented in Figure~\ref{fig:fig 2}. From the figure, We observe that when \(\alpha\) is smaller than \(0.4\), the performance of our approach improves. This indicates that properly fusing local features can enhance the representation ability of features, thereby improving the overall performance. As for the \(k\) value in \textit{top-k} mean pooling, the highest mAP scores are achieved when \(k=32\). We analyze that when \(k\) is too small, the local prediction scores become sensitive to noise, leading to instability in the results. On the other hand, when \(k\) is too large, the local prediction score lacks fine-grained information, as it averages over too many patches, diluting the importance of the most relevant ones. Thus, setting \(k=32\) strikes a balance between these factors, providing the best performance.

\textbf{3) Components Analysis of HP-DVAL.}
To evaluate the contributions of various components to our method, we conduct a series of ablation studies, with the results summarized in Table~\ref{tab 5}. Our baseline experiment uses zero-shot CLIP \cite{pmlr-v139-radford21a}. Our HP-DVAL framework comprises two key stages: Dual-View Alignment Learning (DVAL) and Hierarchical Prompt Tuning (HPT). DVAL leverages CLIP's image encoder to learn a ViT backbone with excellent feature representation ability, while HPT introduces hierarchical prompts to capture task-specific and context-related prior knowledge. Adding these two stages leads to significant improvements across head, medium, and tail classes. The overall mAP outperforms the baseline by 8.01\% on VOC-LT and 16.02\% on COCO-LT. We attribute these gains to the vision-language knowledge transfer guided by this two stages, which can effectively transfer multi-modal knowledge and leverage the powerful feature representation capability of CLIP to tackle CI-MLIC tasks. Integrating HPT with the Semantic Consistency (SC) loss further improves mAP performance, as SC loss can effectively prevent learnable prompts from forgetting the general knowledge encoded in CLIP. \textcolor{red}{Additionally, we compared the results of joint training and two-stage sequential training. Our analysis reveals that joint training impairs the model's ability to focus on distinct optimization objectives, resulting in mixed feature representations and confusion in learning.}

\begin{figure}[t]
\centering
    \begin{minipage}[]{0.72\linewidth}
        \centering
        \includegraphics[width=\textwidth]{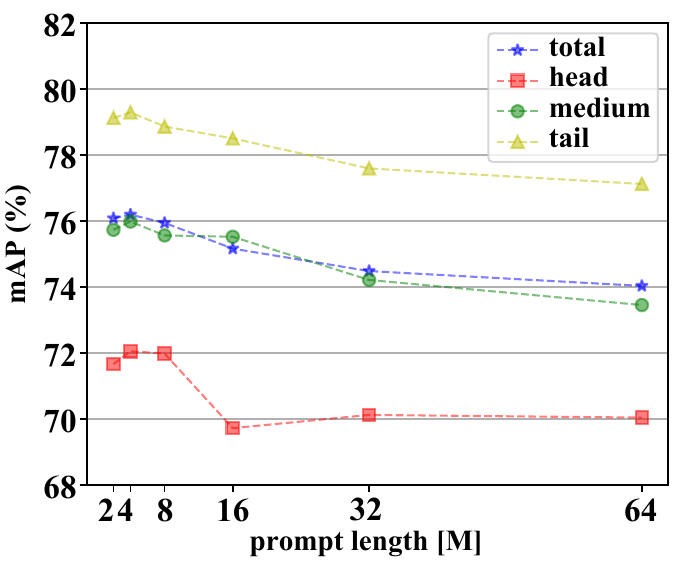}
        \centerline{(a) COCO-LT}
    \end{minipage}
    \begin{minipage}[]{0.72\linewidth}
        \centering
        \includegraphics[width=\textwidth]{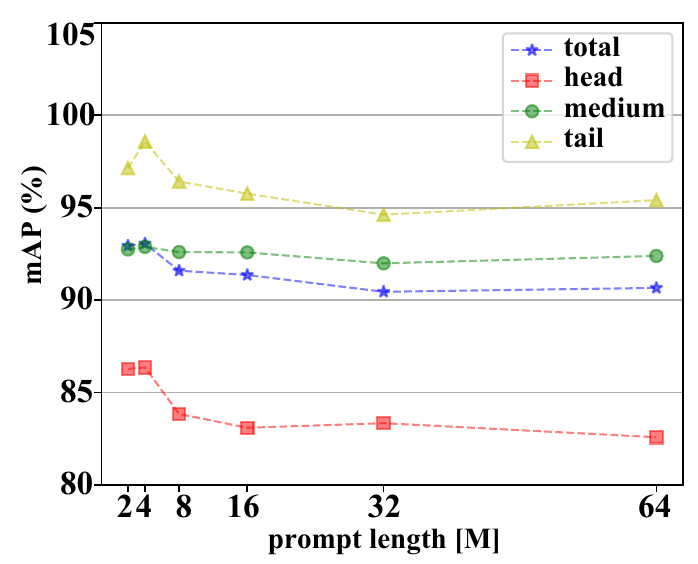}
        \centerline{(b) VOC-LT}
    \end{minipage}
    \caption{\textcolor{red}{The mAP (\%) results with different prompt length \(M\) on COCO-LT and VOC-LT datasets. For both two datasets on long-tailed multi-label classification task, HP-DVAL performs well when \(M\) is small, such as \(2,4\).}}
    \label{fig:fig 5}
\end{figure}

\begin{figure}[t]
	\vspace{-0.1cm}
	\centering
	\includegraphics[width=0.9\linewidth]{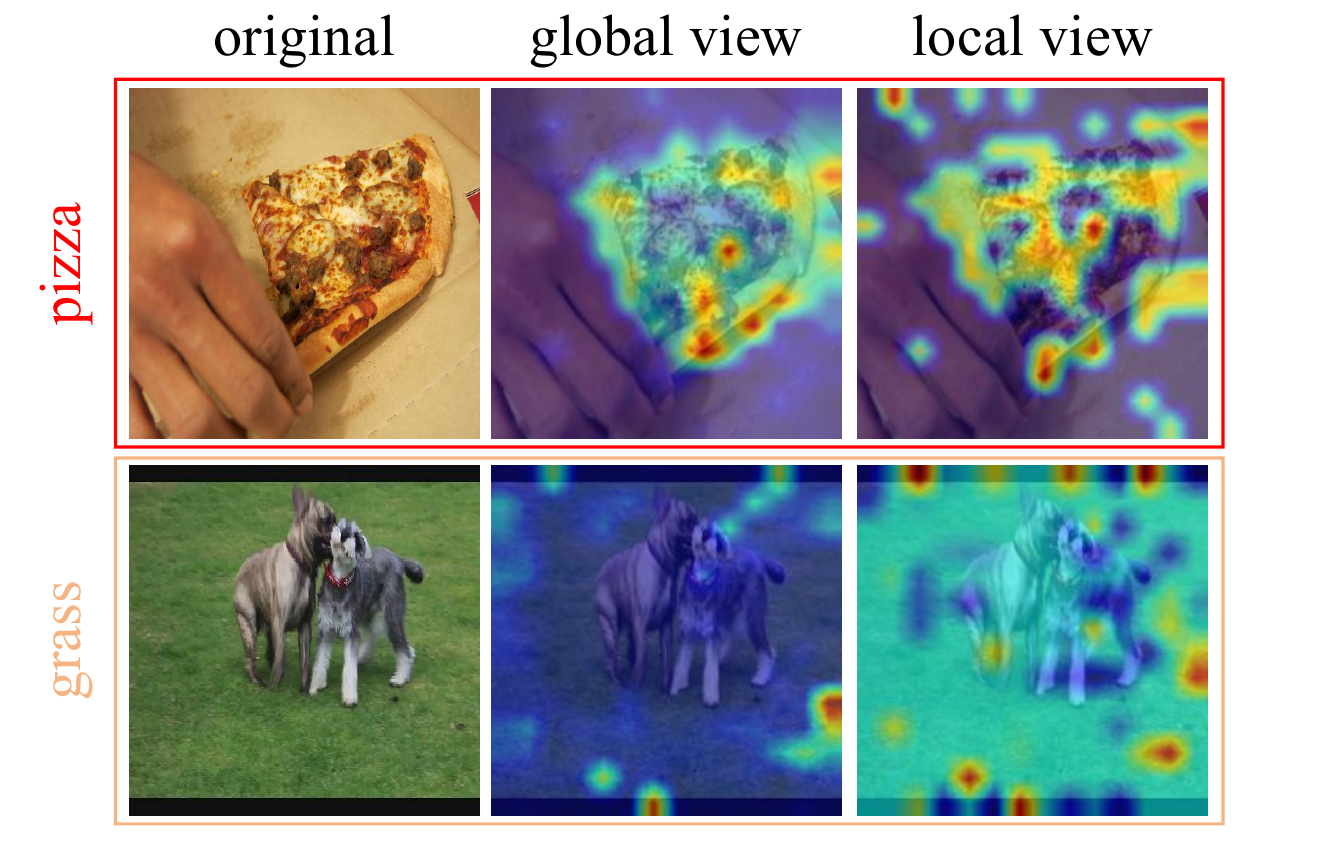}
	\vspace{-0.2cm}
	\caption{\textcolor{red}{Comparative visualization of attention maps from global and local perspectives for "pizza" and "grass".}}
	\label{fig:fig 6}
	\vspace{-0.2cm}
\end{figure}

\textbf{4) Effect of Hierarchical Prompt-Tuning.}
To demonstrate the effectiveness of our proposed hierarchical prompt-tuning, we conduct ablation studies of both global and local prompts. Table~\ref{tab 6} shows the results on VOC-LT and COCO-LT for long-tailed multi-label image classification task. From the table, we can observe that using only one type of learnable prompt will result in performance degradation, whether it is global or local prompts. Integrating two types of prompts can greatly improve performance, which fully demonstrates the superiority of our hierarchical prompts.

\textcolor{red}{To further validate the effectiveness of our method, we visualized the role of these two types of prompts in the image localization process. As shown in Figure~\ref{fig:fig 6}, when the target is prominent (e.g., "pizza"), both global and local prompts can capture high-activation regions. However, when the target category is less salient (e.g., "grass"), our local prompts can still effectively highlight the corresponding region. Overall, the complementary interaction between these two prompts enhances the model's ability to localize specific categories.}

\textbf{5) Effect of Prompt Length.}
As shown in Figure~\ref{fig:fig 5}, we have provided the comparison of the performance of HP-DVAL with different prompt context lengths (\(i.e., M= 2,4,8,16,32,64\)) on the COCO-LT and VOC-LT datasets. For long-tailed multi-label classification, we learn class-shared prompts and thus HP-DVAL performs well when \(M\) is small, such as \(2,4\).

\begin{table}[t]
\centering
\caption{The ablation analysis on hierarchical prompt-tuning of the proposed HP-DVAL on VOC-LT and COCO-LT dataset. No ``\(\surd\)'' means that we use the fixed prompts instead of learnable prompts.}
   \label{tab 6}
\begin{tabular}{cccccc}
\hline
\multirow{2}{*}{Global} & \multicolumn{1}{c|}{\multirow{2}{*}{Local}} & \multicolumn{4}{c}{VOC-LT}                                                                                                       \\ \cline{3-6} 
                        & \multicolumn{1}{c|}{}                       & \multicolumn{1}{c|}{total}          & \multicolumn{1}{c|}{head}           & \multicolumn{1}{c|}{medium}         & tail           \\ \hline
                        & \multicolumn{1}{c|}{\(\surd\)}                       & \multicolumn{1}{c|}{91.36}               & \multicolumn{1}{c|}{83.09}               & \multicolumn{1}{c|}{93.78}               &  95.76              \\
                        \(\surd\)& \multicolumn{1}{c|}{}                       & \multicolumn{1}{c|}{91.59}               & \multicolumn{1}{c|}{83.84}               & \multicolumn{1}{c|}{92.89}               &      96.42          \\
                        \(\surd\) & \multicolumn{1}{c|}{\(\surd\)}                       & \multicolumn{1}{c|}{\textbf{93.06}} & \multicolumn{1}{c|}{\textbf{86.36}} & \multicolumn{1}{c|}{\textbf{92.83}} & \textbf{98.58} \\ \hline
                        &                                             &                                     &                                     &                                     &                \\ \hline
\multirow{2}{*}{Global} & \multicolumn{1}{c|}{\multirow{2}{*}{Local}} & \multicolumn{4}{c}{COCO-LT}                                                                                                      \\ \cline{3-6} 
                        & \multicolumn{1}{c|}{}                       & \multicolumn{1}{c|}{total}          & \multicolumn{1}{c|}{head}           & \multicolumn{1}{c|}{medium}         & tail           \\ \hline
                        & \multicolumn{1}{c|}{\(\surd\)}                       & \multicolumn{1}{c|}{74.03}          & \multicolumn{1}{c|}{70.04}          & \multicolumn{1}{c|}{73.45}          & 77.12          \\
                        \(\surd\) & \multicolumn{1}{c|}{}                       & \multicolumn{1}{c|}{74.48}          & \multicolumn{1}{c|}{70.12}          & \multicolumn{1}{c|}{74.21}          & 77.59          \\
                        \(\surd\) & \multicolumn{1}{c|}{\(\surd\)}                       & \multicolumn{1}{c|}{\textbf{76.19}} & \multicolumn{1}{c|}{\textbf{72.05}} & \multicolumn{1}{c|}{\textbf{75.98}} & \textbf{79.12} \\ \hline
\end{tabular}
\end{table}

\begin{figure}[t]
	\centering
	\includegraphics[width=0.9\linewidth]{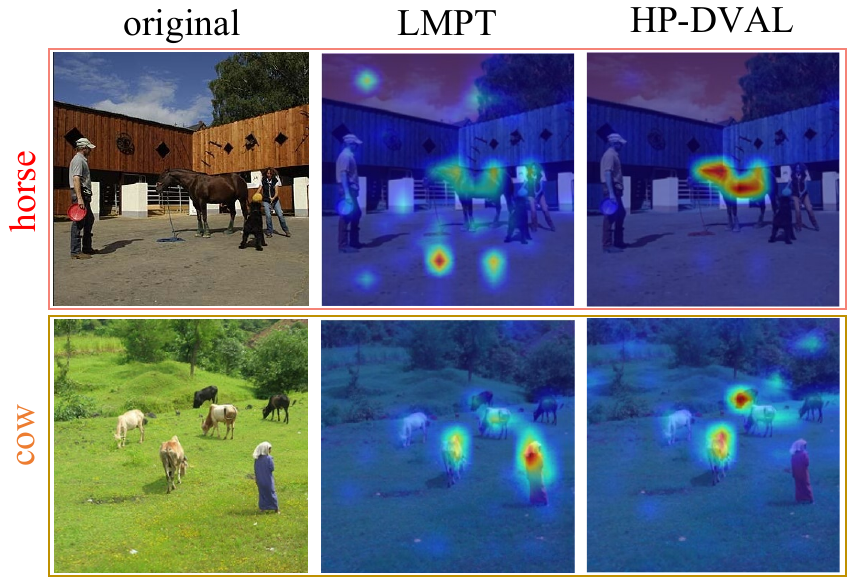}
	\caption{Visualization comparisons of attention maps between LMPT and our method. LMPT pays attention to more irrelevant areas, whereas our method can capture relevant regions more precisely, such as the [horse] region.}
	\label{fig:fig 7}
\end{figure}

\subsection{Qualitative Analysis}
\textcolor{red}{To assess the computational complexity of existing methods, we report three metrics in Table~\ref{tab 7}: model parameters (in millions, M), FLOPs (in GFLOPs), and inference time per image (in ms). Our method shows a slight increase in these metrics, but achieves significantly improved recognition performance (mAP), due to the extra computation introduced by local image patches and prompt tokens. This favorable trade-off between complexity and performance further validates the effectiveness of our design.}

To validate the effectiveness of our proposed method, we conduct visualization experiments. Figure~\ref{fig:fig 7} illustrates the visualization comparisons of attention maps between LMPT and our method. The results show that our method can capture relevant regions more precisely. For instance, in the first column, LMPT pays attention to more irrelevant areas, whereas our method accurately focuses on the [horse] region.

\textcolor{red}{However, while our method employs a semantic consistency (SC) loss to constrain prompts and mitigate overfitting on tail categories, its core design primarily focuses on enhancing feature alignment and fine-grained matching through global-local prompt division, without dedicated mechanisms tailored for extreme class imbalance. This may limit performance in severely imbalanced scenarios, where local prompts may fail to capture distinctive features of tail categories due to insufficient sample support, thereby impairing recognition accuracy. In this regard, exploring differentiated prompt designs for underrepresented classes, as well as leveraging generative models or data augmentation to enrich training data, is worth further investigation to alleviate sample scarcity and improve generalization.}

\begin{table}[t]
	\centering
	\caption{\textcolor{red}{Comparison of computational complexity across methods in terms of model parameters (M), FLOPs (G), and inference speed per image (ms).}}
	\label{tab 7}
	\begin{tabular}{c|c|c|c|c}
		\hline
		Method & Param & FLOPs & Speed & mAP \\ \hline
		CLIP \cite{pmlr-v139-radford21a}   & 149.62         & 19.55     & 11.73      & 60.17   \\
		CoOp \cite{zhou2022learning}   & 149.63         & 20.16     & 11.83      & 60.68   \\
		CoCoOp \cite{Zhou_2022_CVPR} & 150.16         & 20.36     & 11.81      & 61.49   \\
		HP-DVAL   & 150.28         & 20.62     & 14.21      & \textbf{76.19}   \\ \hline
	\end{tabular}%
\end{table}

\section{Conclusion}
\textcolor{red}{
We propose a novel image-text alignment method, termed Dual-View Alignment Learning with Hierarchical-Prompt (HP-DVAL), to address class imbalance multi-label image classification.
By integrating knowledge distillation from VLP models and a hierarchical prompt-tuning strategy, HP-DVAL effectively transfers pretrained vision-language knowledge to improve feature alignment and contextual understanding under imbalanced data distributions. However, the current design lacks dedicated mechanisms for extreme class imbalance, and local prompts may struggle to learn discriminative features for tail categories with limited samples. As future work, we will explore differentiated prompting for rare classes and leverage generative models or data augmentation to enrich training data, thereby improving generalization on underrepresented categories.}

\normalem
\bibliographystyle{IEEEtran}

\bibliography{IEEEtran}



\begin{IEEEbiography}[{\includegraphics[width=0.95in,height=1.25in,clip,keepaspectratio]{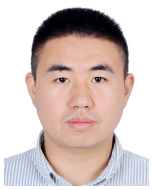}}]{Sheng Huang}
(Member, IEEE) received the B.Eng. and Ph.D. degrees from Chongqing University, Chongqing, China, in 2010 and 2015, respectively. He was a Visiting Ph.D. Student with the Department of Computer Science, Rutgers University, New Brunswick, NJ, USA, from 2012 to 2014. He is currently a Professor with the School of Big Data and Software Engineering, Chongqing University. He has authored/coauthored more than 30 scientific articles in venues, such as CVPR, AAAI, IEEE TRANSACTIONS ON IMAGE PROCESSING, IEEE TRANSACTIONS ON INFORMATION FORENSICS AND SECURITY, and IEEE TRANSACTIONS ON CIRCUITS AND SYSTEMS FOR VIDEO TECHNOLOGY. His research interests include computer vision, machine learning, image processing, and artificial intelligent applications.
\end{IEEEbiography}
\vspace{-0.2cm}
\begin{IEEEbiography}[{\includegraphics[width=0.95in,height=1.25in,clip,keepaspectratio]{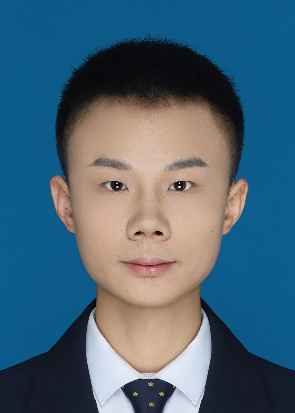}}]{Jiexuan Yan}
is currently pursuing the master’s degree in big data and software engineering from Chongqing University, Chongqing, China. His research interests include computer vision, multi-label image classification, and artificial intelligent applications.
\end{IEEEbiography}
\vspace{-0.2cm}
\begin{IEEEbiography}[{\includegraphics[width=0.95in,height=1.25in,clip,keepaspectratio]{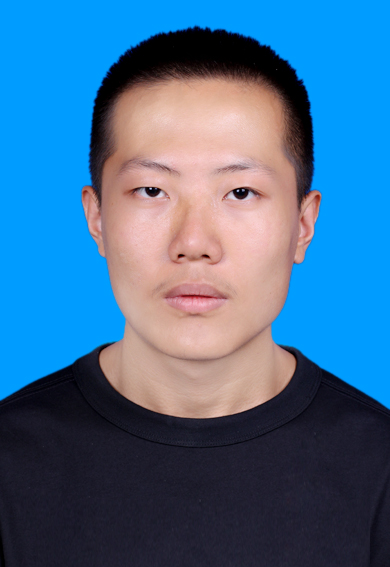}}]{Beiyan Liu}
is currently pursuing the master’s degree in big data and software engineering from Chongqing University, Chongqing, China. His research interests include computer vision, multi-label image classification, and artificial intelligent applications.
\end{IEEEbiography}
\vspace{-0.2cm}
\begin{IEEEbiography}[{\includegraphics[width=0.95in,height=1.25in,clip,keepaspectratio]{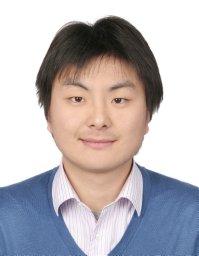}}]{Bo Liu}
received the Ph.D. degree in computer science from Rutgers, The State University of New Jersey, in 2018. He currently is a faculty member at Hefei University of Technology, Hefei, China. Prior to this role, He held the positions of Staff Data Scientist and Senior Research Scientist at Walmart Global Tech and JD.com Silicon Valley Research Center, respectively. His areas of expertise and research interests encompass machine learning, computer vision, and data science.
\end{IEEEbiography}
\vspace{-0.2cm}

\begin{IEEEbiography}[{\includegraphics[width=1.25in,height=1.25in,clip,keepaspectratio]{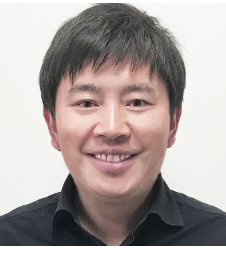}}]{Richang Hong}(Senior Member, IEEE)
is a professor and PhD supervisor at Hefei University of Technology. He currently serves as the Dean of the School of Computer and Information (School of Artificial Intelligence) and the School of Software at Hefei University of Technology. He is also the Deputy Director of the Data Space Research Institute at the Hefei Comprehensive National Science Center and the President of the Anhui Artificial Intelligence Society. His research focuses on artificial intelligence-related fields, with over 300 high-level papers published and more than 20,000 citations. He serves as a Steering Committee member of the International Conference on Multimedia Modeling and as an editorial board member of six international journals, including IEEE TBD, IEEE TMM, and ACM TOMM. He has led various national projects, including the Ministry of Science and Technology’s 863 Program, the Ministry’s Key Research and Development Program, and projects funded by the National Natural Science Foundation of China, such as the Excellent Young Scientists Fund and key foundation projects. His research achievements have been recognized with the National Natural Science Award (Second Prize, 2015), Anhui Provincial Natural Science Award (First Prize, 2017), and Anhui Provincial Science and Technology Progress Award (First Prize, 2020). In teaching, he received the Anhui Provincial Teaching Achievement Award (First Prize, 2022) and the National Teaching Achievement Award (First Prize for Postgraduate Education, 2023). He was awarded the Anhui Province Youth May Fourth Medal in 2019 and was selected for the national leading talent program.
\end{IEEEbiography}

\vfill

\end{document}